\documentclass[review]{elsarticle}

\usepackage{lineno,hyperref}
\usepackage[utf8]{inputenc} % allow utf-8 input
\usepackage[T1]{fontenc}    % use 8-bit T1 fonts
\usepackage{lmodern}
\usepackage{url}            % simple URL typesetting
\usepackage{booktabs}       % professional-quality tables
\usepackage{amsfonts}       % blackboard math symbols
\usepackage{nicefrac}       % compact symbols for 1/2, etc.
\usepackage{microtype}      % microtypography
\usepackage{longtable}      % for long tables
\usepackage{graphicx}
\usepackage{amsmath}
\usepackage{booktabs}
\usepackage{algorithm}
\usepackage{algorithmic}
\usepackage{xcolor}
\definecolor{boston}{rgb}{0.8, 0.0, 0.0}
\usepackage{multirow}
\usepackage{amsfonts}
\usepackage{mathrsfs}  
\usepackage{bm}
\usepackage{setspace}
\usepackage{subcaption}
\usepackage{makecell}
\usepackage{mathtools}
\usepackage{graphicx}
\usepackage[margin=3.0cm]{geometry}
\usepackage{xcolor}
\usepackage{tikz}

\usepackage{makecell}

\definecolor{brightpink}{rgb}{1.0, 0.0, 0.5}

\definecolor{amc}{rgb}{0.0, 0.5  , 0.4}

\newtheorem{theorem}{Theorem}

\newcommand*\linenomathpatch[1]{%
  \cspreto{#1}{\linenomath}%
  \cspreto{#1*}{\linenomath}%
  \csappto{end#1}{\endlinenomath}%
  \csappto{end#1*}{\endlinenomath}%
}
\linenomathpatch{Eq.}
\linenomathpatch{gather}
\linenomathpatch{multline}
\linenomathpatch{align}
\linenomathpatch{alignat}
\linenomathpatch{flalign}

\makeatletter
\newcommand{\heading}[1]% #1 = text
{\par\vskip 1.5ex \@plus .2ex
  \hangindent=1em
  \noindent\makebox[1em][l]{$\,\bullet$}\textit{#1}%
  \par\vskip 1.5ex \@plus .2ex
  \@afterheading}
\makeatother

\begin{document}
	\begin{frontmatter}
		
		\title{\textbf{Supervised Deep Multimodal Matrix Factorization\\for Interpretable Brain Network Analysis}}
		
        \author[1]{Amjad Seyedi\corref{corauthor}}
		\ead{seyedamjad.seyedi@umons.ac.be}

        \author[2]{Lifang He}
        \ead{lih319@lehigh.edu}

        \author[2]{Songlin Zhao}
        \ead{soz223@lehigh.edu}

        \author[3]{Akwum Onwunta}
        \ead{ako221@lehigh.edu}
        
		\author[1]{Nicolas Gillis}
		\ead{nicolas.gillis@umons.ac.be}

        \address[1]{Department of Mathematics and Operational Research, University of Mons, Mons, Belgium}
        \address[2]{Department of Computer Science and Engineering, Lehigh University, Bethlehem, PA, USA}
        \address[3]{Department of Industrial and Systems Engineering, Lehigh University, Bethlehem, PA, USA}
		\cortext[corauthor]{Corresponding author}
		
\begin{abstract}
Multimodal brain network analysis faces a persistent trade-off between predictive accuracy and interpretability. Deep neural networks achieve high accuracy but behave as black boxes that reveal little about the brain modules driving their decisions, whereas matrix factorization methods provide parts-based interpretability yet remain largely shallow, unsupervised, and restricted to a single view, integrating modalities through predefined or heuristic fusion rules. To bridge this gap with a formulation that couples hierarchical modeling capacity with structured, interpretable representations and data-driven fusion, we present Supervised Deep Multimodal Matrix Factorization (SD3MF), an interpretable framework for integrative brain network analysis that generalizes Symmetric Nonnegative Matrix Tri-Factorization (SNMTF) from unsupervised single-graph clustering to supervised prediction over populations of multimodal graphs. SD3MF learns deep hierarchical factorizations for each modality together with a shared latent representation that aligns subjects across modalities. An encoder–decoder formulation jointly optimizes graph reconstruction and supervised prediction, while adaptive weights enable data-driven multimodal fusion. By representing each subject through community-level interaction matrices, the model yields interpretable and discriminative features. Experiments on multimodal connectome datasets show that SD3MF consistently outperforms strong deep learning baselines such as {Convolutional Neural Networks and Graph Neural Networks}, while enabling biologically interpretable insights. 
Code for reproducibility is available at: \href{https://github.com/amjadseyedi/SD3MF}{https://github.com/amjadseyedi/SD3MF}.
\end{abstract}

\end{frontmatter}

\section{Introduction}

Modeling the brain as an interconnected network has recently attracted strong interest because it addresses fundamental questions about cognition, information flow, and distributed brain activity~\cite{TANG2023100046}. Network-based methods aid early detection of neurodegenerative disorders such as Alzheimer’s and Parkinson’s diseases, and HIV-dementia~\cite{sun2022temporal,10385864}. In parallel, multimodal brain network analysis—combining {structural Magnetic Resonance Imaging (sMRI), Diffusion Tensor Imaging (DTI), and functional Magnetic Resonance Imaging (fMRI)}—has emerged as a powerful framework for studying complex brain processes~\cite{tulay2019multimodal,liu2015multimodal}. Integrating these modalities provides a more complete view of network architecture and pathology, supporting improved diagnosis and biomarker discovery.
{
Mathematically, given $N$ subjects represented by $M$ brain-network modalities, let $\mathbf{A}_i^{(m)} \in \mathbb{R}^{n \times n}$ denote the connectivity matrix with $n$ brain regions of subject $i \in \{1,2,\dots, N\}$ in modality $m \in \{1,2,\dots, M\}$, and let $y_i \in \mathcal{Y}$ denote the corresponding clinical or cognitive label. The primary objective of this paper is to learn  a representation  from the connectivity matrices  $\mathbf{A}_i^{(m)}$'s that enables accurate prediction of $y_i$. Beyond predictive performance, the second objective, which is important in brain-network analysis, is to provide interpretable representations that reveal the latent communities of brain regions and their interaction patterns contributing to the prediction. These two goals will be achieved by our proposed model presented in Section~\ref{sec:proposed}.}

% To address the problem above, multimodal brain network analysis has employed a wide range of machine learning methods, from canonical correlation analysis~\cite{benton2019deep,zhou2023attentive} and graph kernels~\cite{SALIM2020103534} to Tensor Decompositions (TDs)~\cite{9177256,10432933,he2018boosted} and CNNs~\cite{wang2017structural,demir2020clustering,chen2023discriminative,zhang2024generalist}. Despite progress, no model consistently captures multimodal brain-network characteristics. Shallow methods struggle with nonlinear topology, Convolutional Neural Networks (CNNs) may distort graph structure, and many approaches ignore cross-modal correlations by treating modalities uniformly. Graph Convolutional Networks (GCNs) have emerged as a strong alternative for graph-structured multimodal data. Methods such as MVGCN~\cite{zhang2018multi} and MVS-GCN~\cite{WEN2022105239} improve prediction in Parkinson’s disease and autism, respectively, but often rely on prior knowledge to build shared spaces. Masking-based variants~\cite{QU2025103570,zhou2024multi} simplify models yet may miss complex inter-modality dependencies, while hybrid methods like GCN-SVM~\cite{10385487} depend on predefined graphs or KNN constructions, limiting generalization.

{To address the problem above, multimodal brain-network analysis has employed a broad range of machine learning approaches, including canonical correlation analysis and graph kernels~\cite{benton2019deep,zhou2023attentive,SALIM2020103534}, tensor decompositions~\cite{9177256,10432933,he2018boosted}, convolutional neural networks (CNNs)~\cite{wang2017structural,demir2020clustering,chen2023discriminative,zhang2024generalist}, and graph convolutional networks (GCNs)~\cite{zhang2018multi,WEN2022105239,QU2025103570,zhou2024multi,10385487}. These approaches offer complementary strengths for multimodal representation and prediction, but important challenges remain. Shallow and tensor-based methods may have limited ability to capture complex hierarchical structures, CNN-based approaches may not fully preserve the underlying graph topology, and GCN-based models can rely on predefined graph constructions, prior information, or simplified mechanisms for multimodal fusion.}

Despite the high predictive accuracy of CNNs and GNNs in brain network analysis, these models often operate as "black boxes". While they capture complex nonlinear patterns, they provide limited insight into identifiable, biologically meaningful brain modules and can exhibit high variance in small-sample settings typical of neuroimaging data. Conversely, traditional Matrix Factorization (MF) methods are inherently linear and offer strong interpretability through parts-based representations, but lack the capacity to model quadratic interactions and hierarchical structure in brain networks. This creates a fundamental gap between expressive but opaque models and interpretable but limited ones, motivating the need for a framework that combines hierarchical modeling power with structured and interpretable representations. We refer to Section~\ref{sec:related} about related work for more details.  
{
To bridge the gaps described above, we introduce Supervised Deep Multimodal Matrix Factorization (SD3MF), which reformulates multimodal graph classification as a supervised deep matrix tri-factorization problem. Starting from symmetric nonnegative matrix tri-factorization (SNMTF), SD3MF replaces the conventional shallow decomposition with a hierarchical encoder--decoder architecture in which each modality is characterized by a sequence of nonnegative node-to-community mappings, while each subject is summarized through a shared low-dimensional community-interaction matrix. This construction allows the factorization to progressively organize connectivity information across multiple levels of abstraction while retaining an explicit structured representation of the original networks. The subject-level interaction matrices are directly coupled to a classifier, so network reconstruction and label supervision are optimized jointly rather than being treated as separate feature-extraction and prediction stages. For multimodal integration, SD3MF further introduces a learnable mixture mechanism that estimates the contribution of each modality from the data, avoiding fixed fusion coefficients or predefined heuristic rules. The resulting formulation therefore brings deep SNMTF, supervised learning, and adaptive multimodal fusion into a single end-to-end framework. We further develop an optimization scheme for jointly estimating its factorization, fusion, and prediction components and investigate key properties of the resulting model, including the roles of non-negativity and over-parameterization as well as its computational efficiency. Although developed in the context of multimodal brain networks, the formulation naturally applies to other multimodal network-valued data, including social, biomedical, sensor, and knowledge networks. Extensive experiments on several multimodal connectome datasets evaluate the proposed formulation against state-of-the-art graph-learning methods and examine the community, connectivity, and salient-ROI structures recovered by the model.

The main contributions of this work are summarized as follows:
\begin{itemize}
    \item We propose SD3MF, a supervised deep matrix factorization framework that extends SNMTF to a hierarchical encoder--decoder architecture for learning multilevel and discriminative representations of brain networks.
    \item We develop a multimodal extension that jointly learns modality-specific hierarchical representations and a shared subject representation, while adaptively weighting the contribution of each modality for principled data fusion.
    \item We formulate an end-to-end optimization scheme for jointly learning the factorization and prediction components, and analyze key properties of the model, including the effects of non-negativity, over-parameterization, and model efficiency.
    \item We conduct extensive experiments on multimodal brain-network datasets, demonstrating competitive or superior predictive performance against state-of-the-art baselines while providing interpretable community, connectivity, and salient-ROI analyses.
\end{itemize}

The remainder of this paper is organized as follows. Section~\ref{sec:related} reviews related work on multimodal brain network analysis and introduces the methodological background on SNMTF, deep SNMTF, and supervised matrix factorization. Section~~\ref{sec:proposed} presents the proposed SD3MF framework, its optimization procedure, theoretical analysis, and computational efficiency. Section~\ref{sec:experiments} reports the experimental setup, quantitative results, ablation studies, and interpretation of the learned representations. Finally, Section~\ref{sec:conclusion} concludes the paper and outlines promising directions for future work.}

\section{Related Work and Background}\label{sec:related}

{This section reviews the main related work and methodological foundations underlying the proposed framework. We first discuss existing approaches for multimodal brain network analysis, with particular emphasis on tensor-based and graph-based methods. We then introduce symmetric nonnegative matrix tri-factorization (SNMTF), its deep hierarchical extension, and supervised matrix factorization, which constitute the main building blocks of our proposed model.}

\subsection{Related Work}

{
Storing relationships among subjects, imaging modalities, and brain connections naturally leads to tensor representations, since tensors provide a convenient way to preserve interactions across multiple modes without collapsing them into a single matrix. Consequently, tensor-based learning has become an important direction for multimodal brain network analysis. Many methods reduce these high-order structures through decomposition to obtain compact latent representations. MPCA~\cite{4359192}, for example, extends Principal Component Analysis (PCA) to multidimensional arrays and has been used to extract features jointly from multiple imaging sources~\cite{9871118}. MIC~\cite{shao2015clustering} constructs similarity tensors and applies CANDECOMP/PARAFAC (CP) decomposition to derive subject-level profiles. Other approaches employ t-product factorizations with sparsity constraints~\cite{8421595}, tensor splitting for multiview clustering~\cite{10.1145/3132847.3132909}, or structured tensor decompositions~\cite{10.5555/3504035.3504050} to capture shared and complementary patterns across multiple graphs or views.

Recent studies have further explored tensor-based fusion strategies for network-valued neuroimaging data. One common approach is to stack adjacency matrices obtained from different functional network estimators into higher-order tensors and then apply CP or Tucker decompositions to learn low-dimensional shared embeddings~\cite{8421595}. Such formulations can preserve multimodal relationships more effectively than independently processing each connectivity matrix. In a complementary direction, \cite{LI2024127497} introduce a slow-thinking module that employs tensor decompositions (TDs) to construct an explicit knowledge graph, thereby refining multimodal fusion and improving disease identification. These developments illustrate the usefulness of tensor representations for organizing heterogeneous brain connectivity information and extracting compact shared structures across modalities.

Despite these strengths, tensor-based methods also face important limitations. Most decomposition models rely primarily on multilinear projections and therefore have limited ability to represent complex hierarchical organization or highly nonlinear relationships among brain networks. In particular, compressing several modalities into a common tensor factorization may obscure subject-specific connectivity patterns or interactions occurring at different levels of network organization. Moreover, the learned latent components are not always directly optimized for the downstream prediction task. As a result, subtle but clinically relevant relationships across modalities and scales may be lost, potentially limiting both predictive performance and interpretability.

Graph Convolutional Networks (GCNs) have consequently become another major direction for multimodal brain network analysis because they can operate directly on graph-structured connectivity data~\cite{KAWAHARA20171038,kipf2017semisupervisedclassificationgraphconvolutional,NIPS2017_5dd9db5e}. Unlike conventional vector-based models, GCNs exploit neighborhood relationships between nodes and can learn increasingly abstract graph representations through successive message-passing layers. Several extensions have been proposed to accommodate multiple imaging modalities. MVGCN~\cite{zhang2018multi}, for instance, aligns modalities within a shared spatial framework, although it relies on auxiliary information that may not always be available in heterogeneous datasets. Attention-based fusion of multiple GCNs~\cite{kazi2019graph} allows modality-specific representations to be combined adaptively, but often requires flattened intermediate feature representations that increase dimensionality and can obscure the original network structure.

Other multimodal GCN strategies impose cross-modal guidance through shared graph patterns~\cite{10182318}; however, their effectiveness can deteriorate when the underlying modalities are poorly aligned. Late-fusion architectures~\cite{ZHANG2023107328} combine modality-specific representations only at later stages of the network, which can restrict interaction between modalities during early representation learning. Explainable GNN approaches have also been investigated for identifying potential biomarkers~\cite{cui2022interpretable}, although their evaluation is often conducted on relatively small cohorts, making generalization an important concern. Methods that reshape functional connectivity using structural priors~\cite{LIU2022102550} or perform multiview graph alignment~\cite{WEN2022105239} can exploit complementary anatomical information, but they typically depend on fixed templates, consistent modality availability, or accurate correspondence between views. Similarly, rigid anatomical constraints may reduce robustness across heterogeneous datasets~\cite{10385487}. Lightweight masking and region-level models~\cite{zhou2024multi,QU2025103570} provide a simpler route toward interpretability, but may oversimplify complex cross-modal interactions or rely on sufficiently accurate image registration.

More recent approaches attempt to combine the complementary strengths of tensor representations and graph neural networks. Some methods first extract shared multimodal representations through TDs and subsequently use them for graph learning~\cite{li2019graph}. Others construct graph operators through Kronecker sums~\cite{zhang2018tensorgraphconvolutionalneural} or use t-product formulations for dynamic and multilayer graph data~\cite{malik2019tensor,huang2020mr}. Low-rank tensor denoising has also been employed to improve connectivity estimation before downstream learning~\cite{10340610}, while structural priors have been incorporated into early GCN layers followed by attention-based refinement~\cite{10430167}. Adversarial variational autoencoder (VAE) frameworks provide another mechanism for aligning latent subspaces across modalities~\cite{10275112}, and complex-valued tensor GNNs have been proposed to model evolving regional interactions~\cite{10095707}. More recently, TGNet~\cite{kong2024tgnet} integrates TDs with multilayer GCNs to jointly exploit high-order multimodal structure and graph topology for disease classification.

Overall, existing studies demonstrate that tensor models are effective at representing high-order multimodal relationships, while GCN-based approaches provide strong nonlinear modeling capacity for graph-structured data. However, tensor methods are often limited in hierarchical and supervised representation learning, whereas GCN-based models can sacrifice structural transparency and interpretability. This motivates the development of a framework that preserves explicit community-level network structure while simultaneously supporting deep hierarchical representation learning, supervised prediction, and adaptive multimodal fusion.

}

\subsection{Symmetric Nonnegative Matrix Tri-Factorization (SNMTF)}
SNMTF is a proximity-aware MF model that facilitates interpretability through structured nonnegative factors.
Let $\mathbf{A} \in \mathbb{R}^{n \times n}_+$ be a nonnegative symmetric matrix, interpreted as the affinity matrix of a graph over $n$ nodes, e.g., $n$ regions of the brain. 
An important task in graph theory is to identify densely connected subsets of nodes, a.k.a.\ communities~\cite{fortunato2010community,11460800}. Most community detection algorithms work in a single-layer fashion; that is, each node is associated with some communities, and possibly, the communities interact with each other. 
A popular model is SNMTF: 
\begin{equation}\label{eq:nmtfApprox}
 \mathbf{A} \approx \mathbf{WSW}^\top ,    
\end{equation}
where $\mathbf{W} \in \mathbb{R}^{n \times r}_+$ contains the community memberships, that is, $W(i,k) > 0$ if node $i$ belongs to the community $k$, while symmetric $\mathbf{S} \in \mathbb{R}^{r \times r}_+$ indicates the interactions between the communities, that is, $S(k,\ell)$ indicates the strength between community $k$ and $\ell$. This model is closely related to the degree-corrected block model~\cite{karrer2011stochastic} and the mixed membership stochastic block model~\cite{airoldi2008mixed}. 

% It is possible to impose $\mathbf{S} = \mathbf{I}$, in which case the communities do not interact, and we have the so-called symmetric nonnegative matrix factorization: $\mathbf{A} \approx \mathbf{WW}^\top$~\cite{kuang2012symmetric}. 
% More recently, this model has been extended in a multi-layer/deep version~\cite{de2023deep}, as follows 
% \begin{align*}
%     \mathbf{A}    & \approx  \mathbf{W}_1 \mathbf{W}_1^\top \\
%     \mathbf{W}_1  & \approx  \mathbf{W}_2 \mathbf{H}_2 \\ 
%     & \vdots  \\
%     \mathbf{W}_{L-1} & \approx  \mathbf{W}_L \mathbf{H}_L, 
% \end{align*} 
% where $\mathbf{W}_\ell \in \mathbb{R}^{n \times r_\ell}_+$ 
% and $\mathbf{H}_\ell \in \mathbb{R}^{r_\ell \times r_{\ell-1}}_+$. 
% Each factorization $\mathbf{W}_{\ell-1}  \approx \mathbf{W}_\ell \mathbf{H}_\ell$ decomposes the communities into larger ones. For example, with two layers, we obtain at the end 
% \[
% \mathbf{A}    \approx  \mathbf{W}_1 \mathbf{W}_1^\top \approx \mathbf{W}_2 \mathbf{H}_2 \mathbf{H}_2^\top \mathbf{W}_2^\top , 
% \] 
% providing two layers of decompositions: with $r_1$ communities at the first layer, and $r_2$ at the second. 

\subsection{Deep SNMTF}
SNMTF was extended to a deeper framework~\cite{hajiveiseh2024deep}, 
as it was done for standard MFs~\cite{trigeorgis2016deep},  
enabling hierarchical representation learning. 
In this deep setting, the  adjacency matrix is factorized in multiple stages, starting with
\mbox{$\mathbf{A} \approx \mathbf{W}_1 \mathbf{S}_1 \mathbf{W}_1^\top$}, where $\mathbf{W}_1$ represents the membership of nodes to low-level communities and $\mathbf{S}_1$ captures the interactions among these communities. The interaction matrix $\mathbf{S}_1$ is then decomposed as
\mbox{$\mathbf{S}_1 \approx \mathbf{W}_2 \mathbf{S}_2 \mathbf{W}_2^\top$}, continuing recursively until $\mathbf{S}_{L-1} \approx \mathbf{W}_L \mathbf{S}_L \mathbf{W}_L^\top$. 
% In each layer, the symmetric (or asymmetric) matrix $\mathbf S_\ell$ representing the interactions between the communities is further decomposed into smaller communities.
Each layer progressively aggregates smaller communities into higher-level structures, resulting in a hierarchical graph summarization that reflects both local and global connectivity patterns.
%In the deep extensions mentioned above~\cite{hajiveiseh2024deep}, 
The optimization problem is formulated using the Frobenius norm~\cite{trigeorgis2016deep, hajiveiseh2024deep}, %that is, $\| \mathbf{A}  -  \mathbf{W}_1 \mathbf{S}_1 \mathbf{W}_1^\top \|_F^2$ is minimized, and similarly for the other layers, 
solving 
\begin{align}\label{eq:dsnmtf}
\min_{\mathbf{\Psi}=\prod_{l=1}^{L}\mathbf{W}_l,\mathbf{S}_l}
    \left\| \mathbf{A}   - \mathbf\Psi\mathbf{S}_l\mathbf\Psi^\top\right\|_F^2 
    \ \text{s.t.} \  \mathbf{W}_l,\mathbf{S}_l\geq 0,
\end{align}
where  $\mathbf\Psi=\prod_{l=1}^{L}\mathbf{W}_l$ denotes the aggregated membership matrix. 
This formulation preserves the nonnegativity and interpretability of MFs while enabling multi-level community discovery within a unified deep optimization framework.
% \ngc{the optimization model is actually unclear; cf.\ my paper with Pierre on consistent deep models.}

\subsection{Supervised Matrix Factorization}
MF is a classical unsupervised feature extraction framework that learns latent structures in complex datasets. Nonnegative MF (NMF), a popular variant, enables interpretable parts-based representations through nonnegativity constraints~\cite{lee1999learning}. Recently, Supervised MF (SMF)~\cite{lee2023exponentially,10.5555/3692070.3693137} extends  NMF by incorporating label information to learn low-dimensional features and perform classification jointly. By aligning representation learning with predictive tasks, SMF improves the relevance of extracted features and has been applied to tabular, single-modal data in areas like text classification, bioinformatics, and image analysis.
Let $a_i \in \mathbb{R}$ be the activation corresponding to label $y_i$, defined as:
\[
a_i=
\begin{cases}
\mathbf{\beta}^\top \mathbf{W}^\top \mathbf{x}_i + \mathbf{\gamma}^\top \mathbf{x}'_i & \text{for SMF-W,} \\
\mathbf{\beta}^\top \mathbf{h}_i +  \mathbf{\gamma}^\top \mathbf{x}'_i & \text{for SMF-H,}
\end{cases}
\]
where $(\mathbf{\beta}, \mathbf{\gamma})$ are logistic regression coefficients associated with either the `filtered' feature $\mathbf{W}^\top \mathbf{x}_i$ (SMF-W), or the learned code $\mathbf{h}_i$ (SMF-H),
% \todo{why don't we use ${\bf H_i}$ instead of ${\bf h_i}$  for notational consistency in the expression for $a_i$ above? }
% \todo{Amjad: I think it is better to keep $\mathbf{h}_i$ to match the main paper’s notation, where $\mathbf{h}_i$ denotes the $i$-th code vector from $\mathbf{H}$. In our notation, $\mathbf{H}_i$ would instead denote a matrix rather than a vector.}
and side information $\mathbf{x}'_i$. 
Let $Z := (\mathbf{W}, \mathbf{H}, \mathbf{\beta}, \mathbf{\gamma})$ denote the block parameters. To estimate $Z$ from observed data $(\mathbf{x}_i, \mathbf{x}'_i, y_i)$ for $i \in \{1, \dots, N\}$, \cite{10.5555/3692070.3693137} consider 
% \ngc{maybe cite where this model comes from exactly}
the following problem:
\begin{equation}
\min_{\substack{\mathbf{W}, \mathbf{H}, \mathbf{\beta}, \mathbf{\Gamma}}}
f(Z) := \xi \| \mathbf{X} - \mathbf{W} \mathbf{H} \|_F^2  + \textstyle \sum_{i=1}^{N} \ell(y_i, a_i)
\end{equation}
with the logistic loss $\ell(y_i, a_i) = \log\left(1 + \exp(a_i)\right) - y_i a_i$. 
SMF provides a principled framework for supervised representation learning that retains the transparency and interpretability of MF while achieving competitive predictive performance. A key weakness is the heuristic synchronization of~\cite{10.5555/3692070.3693137} that separately updates representation and classification components. %which complicates optimization.

% \ngc{If we keep this section, we should explain a bit more, and also highlight the interpretability benefits?}

\section{Proposed Model}\label{sec:proposed}

We propose a supervised graph representation learning framework designed for populations of weighted, undirected/directed networks with a shared node set, where each graph corresponds to an individual subject and the learning goal is graph-level prediction (graph classification). The model is applicable to a broad class of network-valued data in which meaningful low-dimensional structure can be expressed through latent groups of nodes and their interaction patterns. We focus on brain connectomes as a primary application, where nodes represent regions of interest (ROIs) and edges encode functional or structural connectivity. Our key assumption is that such networks can be compactly and discriminatively characterized by a hierarchy of latent communities and by the patterns of intra- and inter-community interactions. To this end, the proposed model learns a shared, possibly deep, node-to-community embedding that captures population-level structure, together with subject-specific community interactions that summarize each network at the community level. These low-dimensional representations are both interpretable and well-suited for supervised learning.  %enabling joint optimization of network reconstruction and prediction performance. 
% \ngc{I think it is a bit hard to understand the construction of the model.}
% We posit that brain networks can be discriminatively characterized by a small number of {functional communities} (clusters of ROIs) and by the patterns of {intra-} and {inter-community} interactions. \ngc{Maybe be a bit more general: what type of data our model can handle? and then mention brain networks is a partiuclar application case on which we will focus in this paper.} 
% Accordingly, our approach learns (i) a shared community-assignment/embedding matrix that groups ROIs into \(r\) communities\ngc{Don't we learn several layers of communities? the $r$ ones are only at the last layer, right?}, and (ii) a subject-specific community interaction matrix that summarizes connectivity between these communities. The resulting low-dimensional interaction matrices provide interpretable, graph-level features for supervised subject classification. 
Let $\{\mathbf{A}_i\}_{i=1}^{N}$  
% \ngc{There could a confusion in the notation here. I think it would be better to keep it consistent. In the previous section, $n$ is the number of nodes, and $L$ the number of layers in the factorization. Here $N$ is the number of graphs and $n$ the number of nodes... could be rather confusing. In fact, you write later that $\mathbf{\Psi}=\prod_{l=1}^{p} \mathbf{W}_l$ so this is not correct. Maybe use $L$ for number of layers in the factorization?}
denote a collection of adjacency matrices (connectomes), where each $\mathbf{A}_i \in \mathbb{R}^{n \times n}$ corresponds to subject $i$  over $n$ ROIs. Associated with each subject is a label $y_i \in \mathcal{Y}$, where $\mathcal{Y}$ denotes the label space (e.g., diagnostic categories or continuous clinical scores). A graph representation function \(f_{\theta}\) maps \(\mathbf{A}_i\) to a compact embedding \(\mathbf{S}_i = f_{\theta}(\mathbf{A}_i)\), and a classifier \(g_{\phi}\) produces the prediction \(\hat{y}_i = g_{\phi}(\mathbf{S}_i)\). In our framework, \(f_{\theta}\) is implemented via (deep) SNMTF, where \(\mathbf{S}_i \in \mathbb{R}^{r \times r}\) explicitly encodes community-to-community interactions.

\subsection{Network Representation Learning}

To extract network-level representations as subject-specific community interactions, we employ Collective Deep SNMTF. 
The key quantity \(\mathbf{S}_i\) summarizes both within-community cohesion (diagonal entries) and between-community coupling (off-diagonal entries). When \(\mathbf{S}_i\) is allowed to take signed values, positive entries indicate strong connectivity (positive correlation) between communities, whereas negative entries indicate anti-connectivity (negative correlation) \cite{hajiveiseh2024deep}. 
To capture shared structure across subjects, we learn a common node-to-community mapping \(\mathbf{\Psi} \in \mathbb{R}^{n \times r}\) by solving 
\begin{align}
\label{eq:cnmtf} 
\min_{\mathbf{\Psi},\mathbf{S}_i}
    \textstyle \sum_{i=1}^{N}
    \left\| \mathbf{A}_i   - \mathbf\Psi\mathbf{S}_i\mathbf\Psi^\top\right\|_F^2 
    \quad \text{s.t.} \quad  \mathbf{\Psi}=\prod_{l=1}^{L} \mathbf{W}_l,\mathbf{W}_l\geq 0,
\end{align} 
% \ngc{not 100\% clear what is nonnegative.}
where $\mathbf{S}_i \in \mathbb{R}^{r \times r}$  is the subject-specific  representation for graph $\mathbf{A}_i$.  
We enforce nonnegativity only on the latent factors $\mathbf{W}_l$, which preserves the interpretability. 
In contrast, $\mathbf{S}_i$ is left unconstrained and may take signed values, to represent both positive and negative interactions. 
% \ngc{This is not the same notation as in the previous section, where you used $\Psi^{(m)}$.}

\subsection{Basic Model: Supervised Deep Matrix Factorization}

We next integrate supervised learning into SNMTF. The goal is to learn structure-aware model parameter set $\mathbf{\Psi}=\prod_{l=1}^{L} \mathbf{W}_l$, interaction-aware network representation $\mathbf{S}_i$, and classifier parameters \(\mathbf{\beta}\) jointly, so that \(\mathbf{S}_i\) remains faithful to the observed graph while becoming predictive of the label \(y_i\).
Notably, although the final classifier is linear, the underlying representation induces quadratic interactions between node memberships and network connectivity, enabling nonlinear modeling of community-level structure prior to classification.
The proposed model integrates two main objectives: (1)~decoder reconstruction, where the factorization $\mathbf{A}_i \approx \mathbf{\Psi} \mathbf{S}_i \mathbf{\Psi}^\top$ enables the model to capture community structures within each graph; (2)~supervised encoding, where each network $\mathbf{A}_i$ is encoded as $\mathbf{\Psi}^\top \mathbf{A}_i \mathbf{\Psi}$, and a supervised loss is applied to its vectorized form $\text{vec}(\mathbf{\Psi}^\top \mathbf{A}_i \mathbf{\Psi})$, to align the learned representation with class labels $y_i$. We propose to solve 
% \ngc{$S_i$ not nonnegative anymore?}\todo{Amjad: Yes, it is no longer constrained to be nonnegative; this is stated explicitly after Eq. (3), where we explain that $\mathbf{S}_i$ is unconstrained to allow signed interactions between latent components.}
% \ngc{Need to define $\Psi$ again}
\begin{align}\nonumber
    \min_{\mathbf{\Psi}, \mathbf{S}_i, \mathbf{\beta}} 
    \textstyle \sum_{i=1}^{N}&
    \mu\| \mathbf{A}_i\text{$-$}\mathbf{\Psi} \mathbf{S}_i \mathbf{\Psi}^\top \|_F^2 
    + \ell\left(y_i,\mathbf{\beta}^\top\text{vec}(\mathbf{\Psi}^\top \mathbf{A}_i \mathbf{\Psi})\right)  
    % \nonumber\\\nonumber  &
    \;\text{s.t.} \; \mathbf{\Psi}= \textstyle \prod_{l=1}^{L} \mathbf{W}_l, \mathbf{W}_l\geq 0, \mathbf \Psi \mathbf 1 = \mathbf 1,
\end{align}
where $\ell(.)$ denotes the cross-entropy loss and $\mu$ is a regularization hyperparameter that balances the decoding (reconstruction) term and the  predictive accuracy. 
The constraint $\mathbf\Psi \mathbf 1 = \mathbf 1$, where $\mathbf 1$ is the vector of all ones of appropriate dimension, allows to interpret $\mathbf\Psi$ as a fuzzy membership matrix. 
The model %forms a deep, structured representation by learning shared nonnegative basis $\mathbf{W}_l$ and projection $\mathbf{\Psi}$, 
% \ngc{why a stochastic projection? What does that mean? can be removed I think}
 %that 
 hierarchically encodes network structure while preserving interpretability. Joint reconstruction and supervision ensure that the learned representations are structurally faithful and predictive. The matrix $\mathbf{S}_i$ captures subject-specific interactions in the learned latent space, providing a compact summary of individual network variations. 

\subsection{Final Model: Supervised Deep Multimodal Matrix Factorization}

Traditional brain network analysis has largely focused on single-modality data, capturing only isolated aspects of brain organization. To address this limitation, we develop a multimodal version of our model that performs model-level fusion, jointly learning from multiple modalities within a single end-to-end architecture. This approach enables a more comprehensive view of brain networks by leveraging complementary information across data types. The model fuses different (e.g., structural and functional) connectivity data into a shared representation space as follows: 
\vspace{-6pt}
\begin{align}
    &\min_{\substack{\mathbf{\Psi}^{(m)},  \\ \mathbf{S}_i,\mathbf{\beta}, \mathbf \alpha}} 
    \textstyle\sum_{i=1}^{N}\sum_{m=1}^{M}
    \mu\| \mathbf{A}^{(m)}_i - \mathbf{\Psi}^{(m)} \mathbf{S}_i {\mathbf{\Psi}^{(m)}}^\top \|_F^2  
    % \nonumber \\ &
    + 
    % \sum_{i=1}^{N}
    \ell(y_i,\mathbf{\beta}^\top\sum_{m=1}^{M}\alpha^{(m)}\text{vec}({\mathbf{\Psi}^{(m)}}^\top \mathbf{A}^{(m)}_i \mathbf{\Psi}^{(m)})) \label{eq:final} 
    \nonumber \vspace{-12pt} \\    &
    \quad\text{s.t.} \quad \mathbf{W}_l^{(m)}\geq 0, \mathbf \Psi^{(m)} \mathbf 1 = \mathbf 1, \textstyle \sum_{m=1}^{M}\alpha_m=1, 
\end{align}
where $\mathbf{A}^{(m)}_i$ is the graph associated to the $i$th subject in mode $m$, $\mathbf \Psi^{(m)} = \prod_{i=1}^L  \mathbf{W}_l^{(m)}$ is the membership matrix of mode $m$, 
$\mathbf \alpha \in \mathbb{R}^M$  control the contribution of each mode in the final classifier. 
% So the predicted probability is
% \begin{align*} p'_i=\mathbf{\beta}^\top\sum_{m=1}^{M}\alpha^{(m)}\text{vec}({\mathbf{\Psi}^{(m)}}^\top \mathbf{A}^{(m)}_i \mathbf{\Psi}^{(m)}),
% \end{align*}
% where \(p'_i\) denotes the predicted logit for subject \(i\). 
The fused feature vector, $v_i = \sum_{m=1}^{M}\alpha^{(m)}\text{vec}({\mathbf{\Psi}^{(m)}}^\top \mathbf{A}^{(m)}_i \mathbf{\Psi}^{(m)})$, is constructed as a convex combination of modality-specific community-level summaries. 
% , \(\mathrm{vec}({\mathbf{\Psi}^{(m)}}^\top \mathbf{A}^{(m)}_i \mathbf{\Psi}^{(m)})\), with weights \(\alpha^{(m)}\) satisfying \(\alpha^{(m)}\ge 0\) and \(\sum_{m=1}^{M}\alpha^{(m)}=1\). 
Intuitively, \({\mathbf{\Psi}^{(m)}}^\top \mathbf{A}^{(m)}_i \mathbf{\Psi}^{(m)}\in\mathbb{R}^{r\times r}\) aggregates ROI-to-ROI connectivity into community-to-community interactions for modality \(m\), yielding an interpretable low-dimensional representation of intra- and inter-community coupling. The classifier then maps the fused representation to a probability via the logistic link, i.e.,
\(\hat{y}_i=\sigma(\mathbf{\beta}^\top \mathbf v_i)\) for binary classification (or \(\mathrm{softmax}\) in the multi-class setting), and the parameters \(\{\mathbf{\Psi}^{(m)}\}_{m=1}^M\), \(\{\mathbf{S}_i\}_{i=1}^N\), \(\mathbf{\beta}\), and \(\boldsymbol{\alpha}\) are learned jointly by minimizing the multimodal objective in \eqref{eq:final}. 
\begin{figure}[H]
    \centering
    \includegraphics[width=1\linewidth]{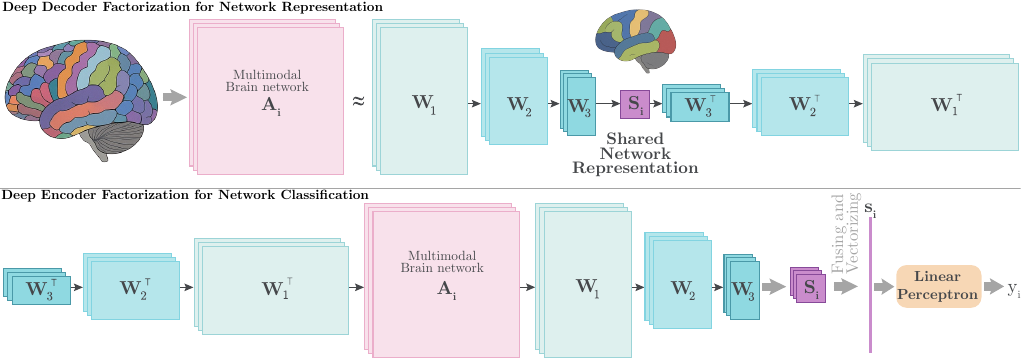}
    \caption{SD3MF architecture: The decoder (top)  %composed of three layers per modality, 
    reconstructs each of the three modality-specific connectomes from a shared subject representation, while the encoder (bottom) produces a discriminative representation  fed to a linear classifier. {Arrows between matrix blocks denote matrix multiplication, while the remaining arrows indicate the processing flow.}}
    \label{fig:model}
\end{figure}
\autoref{fig:model} illustrates the SD3MF pipeline. For each subject \(i\) and modality \(m\), we learn a deep, nonnegative node-to-community mapping parameterized by a product of \(L\) factor matrices, \(\mathbf{\Psi}^{(m)}=\prod_{l=1}^{L}\mathbf{W}^{(m)}_l\). The {decoder} (top) reconstructs the observed connectome via a shared low-rank community interaction matrix \(\mathbf{S}_i\), i.e., \(\mathbf{A}^{(m)}_i \approx \mathbf{\Psi}^{(m)} \mathbf{S}_i {\mathbf{\Psi}^{(m)}}^\top\), which preserves community structure and yields an interpretable representation. 
In parallel, the {encoder} (bottom) produces the community-level summary \({\mathbf{\Psi}^{(m)}}^\top \mathbf{A}^{(m)}_i \mathbf{\Psi}^{(m)} \in \mathbb{R}^{r\times r}\), which is vectorized and optionally fused across modalities using weights \(\{\alpha^{(m)}\}_{m=1}^{M}\). The fused representation is then fed to a linear classifier to predict \(y_i\). 
% {Notably, this community-level projection induces quadratic interactions between node memberships and network connectivity, enabling nonlinear cross-community and cross-modal interactions prior to classification.}
Training jointly optimizes reconstruction and supervised objectives, encouraging \(\mathbf{S}_i\) to remain structurally faithful while being discriminative for classification.

\subsection{Optimization}

The optimization problem is inherently {non-convex} due to the deep MF and shared latent representations across multiple modalities. Rather than relying on explicit regularization, %the model benefits 
we rely on implicit regularization emerging from gradient-based optimization with small learning rates and small-scale initialization, which biases the learned factors toward low-rank and well-conditioned solutions~\cite{arora2019implicit}.
% Each deep basis was initialized with {orthogonal matrices scaled to small magnitudes 
% % \todo{How small? Please rewrite this sentence for clarity}.}
% % \todo{Amjad: I agree and will clarify this by specifying the initialization scale. Detailed settings will be provided in the Appendix (Model Configuration section) for completeness and reproducibility.}
% to stabilize early training and maintain controlled gradient flow. 
Each deep basis was initialized using orthogonal matrices scaled to small magnitudes ($10^{-3}$ in our experiments). Empirically, we observed that small-scale initialization substantially improved optimization stability and final predictive performance, likely by promoting smoother optimization dynamics during the early stages of training.
The shared latent matrices $\mathbf S_i$ were initialized using projections of the input matrices averaged across modalities. %using these initial bases. 
The classifier head was initialized using {Xavier uniform initialization}, also at a small scale, to ensure balanced optimization dynamics across modules.
Training was performed using {stochastic gradient descent (SGD)} with a fixed learning rate of $10^{-5}$ and a maximum of 30{,}000 iterations. 
We specifically favor SGD over adaptive methods like Adam to preserve the gradient noise structure, which is essential for the implicit rank-minimization bias required in matrix factorization~\cite{arora2019implicit}.
The relatively large iteration count allows the small learning rate and implicit regularization dynamics to fully evolve while avoiding instability that could arise from aggressive updates in this highly non-convex setting. Training exhibits stable convergence in practice; a detailed analysis of the optimization behavior is provided in \autoref{app:convergence_theory}.
Additional details, including model efficiency in comparison to GNN-based models, are provided in \autoref{sec:eff}.

\subsection{Optimization Analysis} \label{app:convergence_theory}

In this section, we discuss further our optimization model. 
Collecting all parameters into $\theta = \bigl\{\mathbf W_l^{(m)}\bigr\}_{l,m},\; \{\mathbf S_i\}_{i=1}^N,\; \beta,\; \alpha$, and we denote $\mathbf\Psi^{(m)} = \prod_{l=1}^{L} \mathbf W_l^{(m)}$. 
Given the matrices $\mathbf A_i^{(m)}$'s and the labels $y_i$'s, the SD3MF optimization model is 
\begin{equation}\label{eq:compact}
    \min_{\theta \in \mathcal{C}} \; F(\theta) \;:=\; \underbrace{\sum_{i=1}^{N}\sum_{m=1}^{M} \mu\,\bigl\|\mathbf A_i^{(m)} - \mathbf \Psi^{(m)} \mathbf S_i \mathbf \Psi^{(m)\top}\bigr\|_F^2}_{F_{\mathrm{rec}}(\theta)} \;+\; \underbrace{\sum_{i=1}^{N} \ell\!\bigl(y_i,\;\beta^\top v_i\bigr)}_{F_{\mathrm{cls}}(\theta)},
\end{equation}
where $v_i = \sum_{m=1}^{M}\alpha^{(m)}\text{vec}\Big({\mathbf{\Psi}^{(m)}}^\top \mathbf{A}^{(m)}_i \mathbf{\Psi}^{(m)}\Big)$ is the fused vector,  
and 
the feasible set is 
\[
    \mathcal{C} \;=\; \Bigl\{\theta \;\Big|\; W_l^{(m)} \geq 0,\;\; 
\mathbf{\Psi}^{(m)} = 
    \textstyle\prod_{l=1}^{L} W_l^{(m)}, 
    \mathbf{\Psi}^{(m)} 
    \mathbf{1} = \mathbf{1},\;\; \textstyle\sum_{m=1}^{M}\alpha_m = 1,\;\; \alpha_m \geq 0 \Bigr\}.
\]

This is a non-smooth and highly non-convex problem (both the objective function and the feasible set are non-convex). 
This makes a theoretical convergence analysis of stochastic gradient descent hard and out of the scope of this paper. 
Although~\eqref{eq:compact} is highly non convex, it exhibits partial convexity with respect to some blocks of 
variables, namely the $\mathbf S_i$'s, $\beta$ and $\alpha$. 
This could be leveraged in further study of optimization algorithms and convergence 
behavior. 
In any case, our experimental results show  that the proposed optimization strategy converges on all considered data sets; see Section~\ref{sec:empiricalconv} and \autoref{fig:conv_all}.  In the following, we discuss some of aspects of this optimization model.

\subsubsection{Effect of Non-negativity Constraints}

The constraint $\mathbf \Psi^{(m)} \geq 0$ provides improved identifiability; this follows from the vast literature on NMF~\cite[Chapter~4]{gillis2020NMF}. For example, we have the following results.  
\begin{theorem}[Identifiability under separability, \cite{arora2018learning}]\label{thm:nmf_id} 
Consider $\mathbf A =  \mathbf{\Psi S \Psi}^\top$ 
with $\mathbf \Psi \in \mathbb{R}^{n \times r}_+$,  $\mathrm{rank}(\mathbf S) = r$, and $\mathbf \Psi$ is separable, that is, $\mathbf \Psi$ contains a non-singular $r$-by-$r$ diagonal submatrix. 
Any factorization of $\mathbf A$ of the same form will recover $\mathbf \Psi$ and $\mathbf S$, up to permutations and scalings of the columns of $\mathbf \Psi$ and counter scaling of rows and columns of $\mathbf S$.  
%If $W$ satisfies the \emph{separability} (anchor) condition---i.e., for each community $k$ there exists at least one node $i$ such that $W_{i,k} > 0$ and 
%$W_{i,j} = 0$ for all $j \neq k$---then $W$ is essentially unique up to permutation and scaling of its columns. %In the deep setting, this uniqueness applies to the end-to-end product $\Psi = \prod_{l=1}^{L} W_l$, not to the individual factors $W_l$.
\end{theorem}

In brain network terms, separability means that each community has at least one 
anchor region that belongs exclusively to that community. This is a mild and 
neurobiologically plausible assumption, since many brain regions are strongly 
associated with a single functional network~\cite{yeo2011organization}.
We note that an alternative identifiability route relies on the sufficiently 
scattered condition (which relaxes separability, and requires some degree of sparsity in $\mathbf \Psi$) combined with volume minimization of 
$\mathbf S$~\cite{huang2016anchor, fu2018anchor}; this would be an interesting direction of further research: use volume minimization for relaxed identifiability conditions. 

%however, since SD3MF does not minimize the volume of $S$, we appeal to separability instead.
% \ngc{This is true if you minimize the volume of $S$, which we do not do \cite{huang2016anchor, fu2018anchor}. 
% Rather use Arora paper: it is unique under separability of $W$ (for each community, one brain region only belongs to that community --anchor-region assumption). \cite{arora2018learning}} 

The deep product $\mathbf \Psi^{(m)} = \prod_l \mathbf W_l^{(m)}$ introduces additional factorization ambiguity between layers: the individual $\mathbf W_l^{(m)}$ are not uniquely identifiable~\cite{de2023consistent}. However, the non-negativity of each $\mathbf W_l^{(m)}$ ensures that $\mathbf \Psi^{(m)}$ itself remains nonnegative, so the identifiability of $\mathbf \Psi^{(m)}$ under separability is preserved. This eliminates the sign and rotational ambiguities present in unconstrained deep factorizations at the level of the end-to-end mapping, which is the quantity that determines both the reconstruction and the classifier output.

\subsubsection{Optimization Effects of Over-parameterization}

Deep factorization acts as an over-parameterization that can favorably alter the optimization landscape. For the reconstruction component, define the per-subject loss
\begin{equation}\label{eq:per_subject}
    f_i(\Psi, \mathbf S_i) \;=\; \bigl\|\mathbf A_i - \mathbf \Psi \mathbf S_i \mathbf \Psi^\top\bigr\|_F^2.\nonumber
\end{equation}
With $\mathbf \Psi =\mathbf  W_1 \mathbf W_2 \mathbf W_3$, the product parameterization can alter the curvature and local geometry of the optimization landscape compared with optimizing directly over a single matrix~$\mathbf \Psi$~\cite{de2023consistent, allen2019convergence}. Informally, the product parameterization may lead to a more favorable optimization geometry, potentially facilitating gradient-based optimization.

Note that in principle, the shallow model ($L=1$) has the same expressive power as the deep model for classification, since the classifier operates on $\mathbf \Psi^\top \mathbf A_i \mathbf \Psi$ regardless of whether $\mathbf \Psi$ is parameterized as a single matrix or a product of factors. The consistent empirical improvement of the deep model over the shallow variant, see \autoref{fig:ablation}, therefore provides direct evidence that depth improves optimization rather than expressiveness: the over-parameterized product structure helps gradient descent reach better stationary points by smoothing the loss landscape.

% \ngc{
% In Theory, the shallow model should be able to provide as good results, because deepness is not used in the classification part (but can be used for the interpretability). This shows that deepnees allows to reach better solutions thanks to the overparametrization.  
% }

% \ngc{Mention figure from the paper comparing shallow and deep results?} 

% \subsection{Summary}

% \begin{table}[h]
% \centering
% \caption{Summary of convergence properties.}
% \label{tab:convergence_summary}
% \small
% \begin{tabular}{lll}
% \toprule
% \textbf{Property} & \textbf{Guarantee} & \textbf{Reference} \\
% \midrule
% Stationary-point convergence & $\min_t \mathbb{E}[\|G_\eta(\theta^{(t)})\|^2] \to 0$ & Prop.~\ref{prop:sgd} \\
% Constant step-size regime & $\mathcal{O}(1/T + \eta\sigma^2)$ neighborhood & Remark after Prop.~\ref{prop:sgd} \\
% Implicit low-rank bias & $\Psi \to \arg\min \|\Psi\|_*$ at convergence & Thm.~\ref{thm:implicit}~\cite{arora2019implicit} \\
% NMF identifiability & Essential uniqueness under separability & Thm.~\ref{thm:nmf_id}~\cite{arora2018learning} \\
% Block multi-convexity & $F$ convex in $S_i$, $\beta$, $\alpha$ individually & \S\autoref{app:convergence_theory} \\
% Landscape smoothing & Fewer bad local minima via depth & \cite{de2023consistent, allen2019convergence} \\
% \bottomrule
% \end{tabular}
% \end{table}

These results do not guarantee global optimality---we are dealing with a non-convex problem most likely NP-hard---but they collectively justify the stable convergence observed empirically and confirm that the architectural choices in SD3MF (deep parameterization, small initialization, non-negativity, reconstruction regularization) are theoretically grounded.

\subsection{Model Efficiency}\label{sec:eff}
 
In the deep formulation of SD3MF, each modality is parameterized by a product of three nonnegative factor matrices $\mathbf W_1 \in \mathbb{R}^{n \times r_1}$, $\mathbf W_2 \in \mathbb{R}^{r_1 \times r_2}$, and $\mathbf W_3 \in \mathbb{R}^{r_2 \times r_3}$. With $(r_1,r_2,r_3)=(30,20,10)$, the number of parameters per modality is $n r_1 + r_1 r_2 + r_2 r_3 = 30n + 800$, which scales linearly with the number of nodes and remains independent of graph convolution depth or feature dimensionality. The intermediate ranks $r_1$ and $r_2$ are chosen to enable a gradual hierarchical dimensionality reduction, while the deepest rank $r_3=10$ reflects the assumed number of communities and aligns with evidence that brain functional architecture comprises approximately ten major functional networks \cite{smith2009correspondence}. In addition, the subject-specific interaction matrices contribute only $N r_3^2 = 100N$ parameters, and the final linear classifier introduces a negligible number of additional weights.
In contrast, GNN-based multimodal architectures employ multiple graph convolution layers with weight matrices of size $O(d_{\mathrm{in}} d_{\mathrm{out}})$ per layer, together with parameters for neighborhood aggregation, modality pooling, and fully connected classifiers, resulting in substantially higher parameter counts and memory consumption. Moreover, repeated message-passing operations incur additional runtime and memory overhead proportional to the number of edges and layers. By relying on low-rank matrix factorizations and closed-form projections, SD3MF avoids explicit neighborhood aggregation and yields a compact architecture with favorable parameter efficiency and computational complexity.

\section{Experiments}\label{sec:experiments}

\subsection{Implementation Details}\label{sec:imp}

We implemented SD3MF in PyTorch and optimized all parameters end-to-end using stochastic gradient descent with a learning rate of $10^{-5}$ and a maximum of 30,000 iterations. For each modality, the deep node-to-community mapping $\mathbf \Psi^{(m)}$ was parameterized by a three-layer factorization with hidden widths ([30, 20, 10]), and nonnegativity was enforced through ReLU activations followed by row normalization. All deep factor matrices were initialized using orthogonal initialization 
% (\texttt{torch.nn.init.orthogonal\_}) 
Scaled to small magnitudes ($10^{-3}$) to stabilize early training and induce implicit regularization. The shared interaction matrices $\mathbf S_i$ were initialized from modality-averaged projections using the initial bases and symmetrized at each iteration. Modality weights $\alpha^{(m)}$ were learned via a softmax over trainable logits to ensure a simplex constraint. The linear classifier head was initialized with Xavier initialization at a small gain ($10^{-3}$) to balance optimization dynamics across modules. All experiments were implemented in PyTorch (Python) and executed on a laptop with an NVIDIA RTX 4060 GPU with 8 GB VRAM; the model can also be trained and evaluated on CPU with comparable behavior but longer runtimes.

\subsection{Datasets}\label{sec:datsaets}

We assessed our approach on three real clinical neuroimaging datasets spanning multiple imaging modalities. These datasets are of small to moderate scale, reflecting typical cohort sizes in multimodal brain studies. To limit bias, major confounds (e.g., age and gender) were handled during recruitment and preprocessing so that group distributions remained comparable. Acquisition was also kept consistent within each dataset to support fair comparisons.

\paragraph{Human Immunodeficiency Virus Infection (HIV)}
Data were collected in the Early HIV Infection Study at Northwestern University and include both fMRI and DTI~\cite{doi:10.1212/WNL.0b013e318278b5b4}. We used a standard preprocessing workflow~\cite{cao2015identifying}. The fMRI data were processed using the DPARSF toolbox, including realignment to the first volume, slice-timing correction, normalization to the MNI template, and spatial smoothing with an 8-mm Gaussian kernel. Subject-specific whole-brain networks were then constructed by parcellating the brain into 90 cerebral regions, excluding 26 cerebellar regions, and computing pairwise connectivity using correlation coefficients. The dataset contains 35 healthy and 35 diseased subjects.

\paragraph{Bipolar Disorder (BP)}
This dataset comes from the UCLA Ahmanson-Lovelace Brain Mapping Center and includes 52 euthymic bipolar I participants and 45 age- and gender-matched healthy controls, with both fMRI and DTI modalities~\cite{AJILORE201537}. Resting-state fMRI was acquired on a 3T Siemens Trio scanner using a T2* EPI gradient-echo sequence with IPAT, while DTI was acquired on a Siemens 3T Trio scanner. Brain networks were generated using the functional connectivity (CONN) toolbox~\cite{whitfield2012conn}. Raw EPI data were realigned and co-registered, followed by normalization and smoothing. Nuisance effects from motion, white matter, and cerebrospinal fluid (CSF) were regressed out, and connectivity matrices were computed from pairwise blood-oxygen-level-dependent (BOLD) correlations across 82 FreeSurfer-derived cortical and subcortical gray-matter regions.

\paragraph{Parkinson's Progression Markers Initiative (PPMI)}
This dataset is derived from the Parkinson's Progression Markers Initiative (PPMI), a large collaborative study aimed at improving Parkinson's disease therapeutics. We use DTI data from 754 subjects. The diffusion data were corrected for head motion and eddy-current distortions, skull-stripped, and linearly registered using FreeSurfer following~\cite{zhan2015comparison}. We parcellated 84 ROIs from the T1 images with FreeSurfer and, using these ROIs, constructed three structural connectivity matrices per subject via whole-brain probabilistic tractography: Probabilistic Index of Connectivity (PICo), Hough voting, and Probtrackx~\cite{zhan2015comparison}.

% \begin{table}
% \centering
% \scriptsize
% \caption{Datasets used in the experiments. Each dataset comprises two classes (Healthy vs. Diseased patient) across multiple functional and structural imaging modalities. 
% }
% \label{tab:datasets}
% \begin{tabular}{lccc}
% \hline
% \textbf{Dataset} & \textbf{Feature Size} & \textbf{Modalities} %& \textbf{Class} 
% & \textbf{\# Healthy / Diseased} \\
% \hline
% HIV & 
% 90$\times$90$\times$2 
% & \{fMRI, DTI\} 
% & 35 / 35 \\ % \\ &  & DTI %& Patient  & 35 \\
% BP & %\multirow{2}{*}
% {82$\times$82$\times$2} & \{fMRI, DTI\} 
% %& Healthy 
% & 45 / 52 \\ 
% %\\ &  & DTI & Patient & 45 / 52 \\
% %\hline
%  PPMI & 84$\times$84$\times$3 & %\makecell{PICo\\Hough\\Probtracx} 
%  \{PICo, Hough, Probtracx\}
%  %& \makecell{Healthy\\Patient} 
%  & 149 / 569 \\ %\makecell{149\\569}\\
% \hline
% \end{tabular}
% \end{table}

% {The main characteristics of the three datasets, including the number of subjects, modalities, and brain regions, are summarized in Table~\ref{tab:datasets}.}
For each dataset, we employ a stratified random split with 80\% of the samples used for training and 20\% for evaluation. The procedure is repeated over 10 runs with different random seeds, and we report the mean and standard deviation across runs.

\subsection{Results}
\autoref{tab:results} reports the classification performance of all compared methods, including the same baselines considered in~\cite{kong2024tgnet}, which represent state-of-the-art approaches for supervised and unsupervised graph-based classification. Baseline results are taken from prior work evaluated on the same datasets, while SD3MF is implemented and evaluated using the same preprocessing and experimental protocol to ensure fair comparison. For our model, the architectural and optimization hyperparameters are selected following the protocol described in the \autoref{sec:imp}. Overall, SD3MF consistently achieves the best or near-best performance across all datasets and metrics, showing both superior predictive accuracy and strong ranking capability.

\begin{figure}
    \centering
    \includegraphics[width=0.4\linewidth]{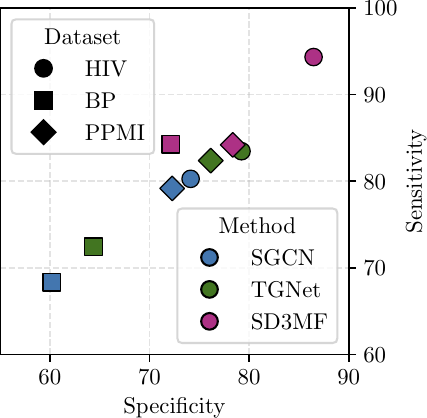}
    \caption{Sensitivity vs. specificity of top methods.}
    \label{fig:sens_spec}
\end{figure}

On HIV, a challenging and high-variance dataset, shallow multimodal methods perform poorly, while graph-based models yield substantial gains, highlighting the importance of relational modeling. TGNet is the strongest baseline (ACC 81.39\%, AUC 82.08\%), but SD3MF significantly outperforms all competitors (ACC 87.50\%, AUC 95.31\%) with markedly lower variance, indicating more effective multimodal integration and improved stability.
On BP, GCN-based methods outperform kernel and CNN baselines, with TGNet again providing the best baseline performance (ACC 67.78\%, AUC 66.31\%). SD3MF achieves a clear margin (ACC 74.00\%, AUC 85.05\%), with a particularly large AUC gain, suggesting more discriminative representations in difficult or imbalanced settings and improved robustness.
PPMI is comparatively less noisy, with higher baseline accuracies across methods. While TGNet attains the highest baseline ACC (83.94\%), SD3MF obtains the best AUC (79.89\%) with competitive accuracy (82.37\%) and the smallest variance among top methods, indicating better calibration and stability on structured data.
Across all datasets, graph-based approaches consistently outperform traditional multimodal methods, confirming the necessity of relational learning. SD3MF further improves both accuracy and robustness, particularly in low-sample regimes common in medical imaging.
To further evaluate clinical performance, \autoref{fig:sens_spec} illustrates the sensitivity–specificity trade-off across top-performing methods and datasets. SD3MF consistently lies in the top-right region, demonstrating a more favorable balance between sensitivity and specificity, which is particularly important in clinical settings where both false positives and false negatives are critical.

\begin{table}
\small
\caption{Performance comparison on HIV, BP, and PPMI datasets.}\label{tab:results}
\begin{tabular}{l|cc|cc|cc}
\hline
\multirow{2}{*}{Method} & \multicolumn{2}{c}{HIV} &\multicolumn{2}{c}{BP} & \multicolumn{2}{c}{PPMI} \\
\cline{2-7}
 & ACC & AUC & ACC & AUC & ACC & AUC\\
\hline
M2E     & 50.61$\pm$15.84 & 51.53$\pm$13.68 & 57.78$\pm$12.61 & 53.63$\pm$11.82 & 76.98$\pm$8.65 & 71.53$\pm$8.34 \\
MIC     & 55.63$\pm$15.28 & 56.61$\pm$13.43 & 51.21$\pm$13.78 & 50.12$\pm$16.78 & 78.03$\pm$8.36 & 72.42$\pm$7.93 \\
MPCA    & 67.24$\pm$11.56 & 66.92$\pm$12.46 & 56.92$\pm$13.33 & 56.86$\pm$13.69 & 81.25$\pm$6.34 & 72.36$\pm$7.18 \\
MK-SVM  & 67.71$\pm$13.18 & 69.89$\pm$10.36 & 60.12$\pm$10.83 & 56.78$\pm$12.86 & 81.68$\pm$5.06 & 75.96$\pm$6.89 \\
% \hline
3D-CNN  & 74.31$\pm$18.81 & 73.53$\pm$16.41 & 63.33$\pm$11.21 & 61.62$\pm$10.26 & 82.24$\pm$5.32 & 75.65$\pm$6.11 \\
GAT     & 68.58$\pm$13.51 & 67.31$\pm$14.32 & 61.31$\pm$15.04 & 59.93$\pm$13.54 & 82.28$\pm$5.95 & 75.19$\pm$5.57 \\
GCN     & 70.16$\pm$12.54 & 69.94$\pm$12.91 & 64.44$\pm$15.71 & 64.24$\pm$16.45 & 83.10$\pm$6.01 & 75.33$\pm$6.79 \\
DiffPool& 71.42$\pm$14.78 & 71.08$\pm$15.12 & 62.22$\pm$12.83 & 62.54$\pm$13.41 & 82.19$\pm$6.23 & 72.75$\pm$2.18 \\
MVGCN   & 74.29$\pm$11.27 & 73.75$\pm$12.63 & 62.22$\pm$16.83 & 62.64$\pm$16.89 & 82.54$\pm$5.15 & 76.48$\pm$6.69 \\
MVS-GCN & 78.57$\pm$16.66 & 77.50$\pm$12.94 & 65.00$\pm$16.03 & 63.70$\pm$15.31 & 80.28$\pm$1.41 & 76.54$\pm$8.03 \\
MaskGNN & 75.71$\pm$11.41 & 79.20$\pm$12.43 & 64.95$\pm$10.34 & 64.71$\pm$10.31 & 80.43$\pm$5.68 & 72.24$\pm$5.14 \\
GCN-SVM & 62.86$\pm$6.55  & 63.33$\pm$15.00 & 61.67$\pm$5.15  & 57.33$\pm$8.21  & 78.87$\pm$4.86 & 70.25$\pm$9.59 \\
SGCN    & 77.14$\pm$14.09 & 80.47$\pm$12.67 & 63.89$\pm$17.24 & 65.18$\pm$16.73 & 81.09$\pm$5.17 & 74.78$\pm$7.69 \\ 
TGNet   & 81.39$\pm$13.41 & 82.08$\pm$14.81 & 67.78$\pm$12.28 & 66.31$\pm$10.24 & \textbf{83.94$\pm$6.09} & 77.93$\pm$6.86 \\
% \hline
% \textbf{S3MF} & &  & &  &  & \\
\textbf{SD3MF} & \textbf{87.50$\pm$3.60} & \textbf{95.31$\pm$2.62} & \textbf{74.00$\pm$4.90} & \textbf{85.05$\pm$2.34} & {82.37$\pm$2.07} & \textbf{79.89$\pm$2.53} \\
\hline
\end{tabular}
\end{table}

\subsection{Ablation Study}
To better understand the behavior of SD3MF, we perform several ablation studies. 

\subsubsection{Effect of decoder factorization} 
\begin{figure}[H]
    \centering

    \begin{subfigure}{0.32\linewidth}
        \centering
        \includegraphics[width=\linewidth]{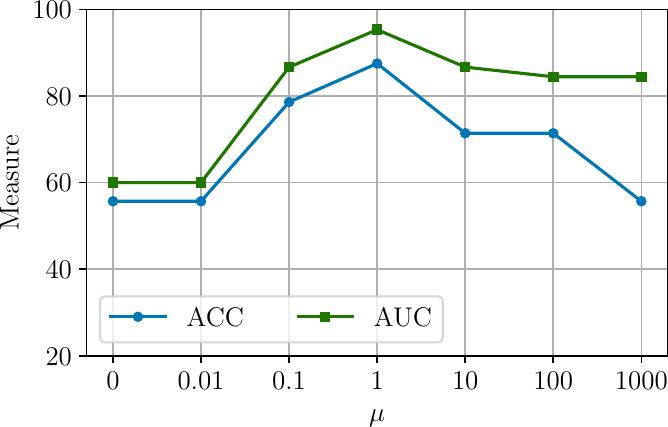}
        \caption{HIV dataset}
    \end{subfigure}
    \hfill
    \begin{subfigure}{0.32\linewidth}
        \centering
        \includegraphics[width=\linewidth]{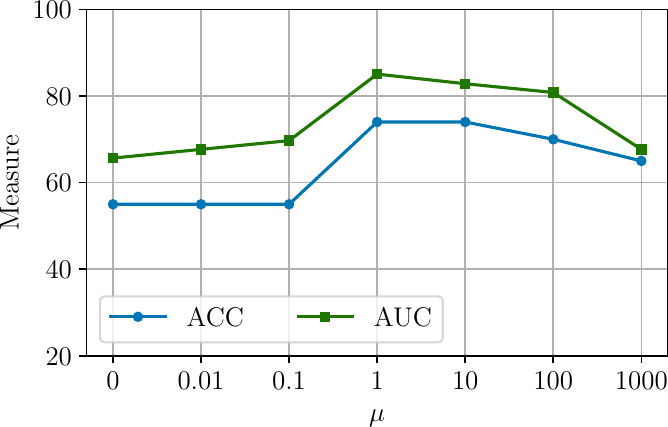}
        \caption{BP dataset}
    \end{subfigure}
    \hfill
    \begin{subfigure}{0.32\linewidth}
        \centering
        \includegraphics[width=\linewidth]{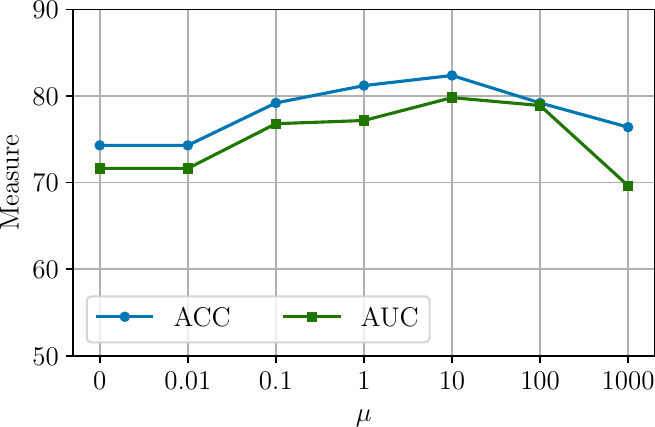}
        \caption{PPMI dataset}
    \end{subfigure}
    \caption{Influence of the regularization parameter $\mu$ on model performance. Both ACC and AUC improve as $\mu$ increases up to $\mu \in [1,10]$ and decrease for larger values.}
    \label{fig:parameter}
\end{figure}

We analyze the influence of the decoder reconstruction term by evaluating different values of the regularization parameter $\mu$ in independent runs. As shown in \autoref{fig:parameter}, at $\mu$=$0$, where the decoder reconstruction term is removed, performance is substantially lower, confirming that decoder factorization plays a crucial role in regularizing the representation and improving generalization. Both ACC and AUC increase steadily as $\mu$ grows from very small values to $\mu$$\in$$[1\ 10]$, indicating that incorporating decoder-based reconstruction promotes the learning of structurally faithful and more discriminative community representations. When $\mu$ becomes too large, performance degrades, suggesting that excessive emphasis on reconstruction weakens the supervised signal and leads to suboptimal classification. This behavior highlights the importance of balancing decoder factorization and supervised encoding: a moderate reconstruction weight enables SD3MF to preserve network structure while enhancing predictive performance. 
% For the HIV and PPMI datasets, the corresponding results are reported in \autoref{sec:mu}.
% \todo{Why the pot for BP dataset only? Can you include the plots and discussions for HIV and PPMI datasets for comparison?} \todo{Amjad: I'll generated those plots for HIV and PPMI and added them (along with a brief discussion) to the Appendix.}
\subsubsection{Effect of Modality Types}
To evaluate the contribution of multimodal fusion, we conduct an ablation study by training and testing the model using each modality independently and comparing the results with the full multimodal configuration (\autoref{tab:modals}). Treating each modality in isolation effectively removes the fusion component of SD3MF, allowing us to isolate the impact of integrating complementary neural connectivity information. 
As reported in \autoref{tab:modals}, SD3MF consistently achieves improved performance when combining modalities across all datasets. On the HIV dataset, the DTI-only variant outperforms the fMRI-only model, suggesting that structural connectivity provides more stable disease-related patterns in this cohort, while the multimodal model further improves performance by incorporating complementary functional information. A similar trend is observed on the BP dataset, where neither modality alone reaches the performance of the fused model, highlighting the importance of cross-modal interactions. For the PPMI dataset, although individual tractography-derived modalities already yield competitive results, their combination leads to the best overall accuracy and AUC, indicating that different tractography methods encode partially distinct yet complementary structural information.
Importantly, these patterns are reflected in the learned modality weights $\alpha^{(m)}$, which adaptively assign higher importance to more informative modalities while still preserving contributions from weaker ones, providing a principled and data-driven fusion mechanism within SD3MF. 
Overall, this ablation study confirms that the performance gains of SD3MF are not solely due to strong single-modal modeling, but critically rely on its adaptive multimodal integration strategy.

\begin{table}[H]
\centering
\caption{Evaluating the diagnostic accuracy of the SD3MF model across single-modal and multimodal neural connectivity data.}
\label{tab:modals}
\begin{tabular}{llccc}
\hline
\textbf{Dataset} & \textbf{Modality} & \textbf{Accuracy (\%)} & \textbf{AUC (\%)} \\
\hline

\multirow{3}{*}{HIV}
& fMRI  & 77.16$\pm$3.22   & 89.58$\pm$4.89 \\
& DTI   & 82.13$\pm$6.84   & 91.15$\pm$2.62 \\
& Both  & 87.50$\pm$3.60   & 95.31$\pm$2.62 \\
\hline

\multirow{3}{*}{BP}
& fMRI  & 71.00$\pm$5.50   & 78.18$\pm$2.50 \\
& DTI   & 72.00$\pm$2.70   & 76.57$\pm$2.66 \\
& Both  & 74.00$\pm$4.90   & 85.05$\pm$2.34 \\
\hline

\multirow{4}{*}{PPMI}
& PICo       & 78.93$\pm$2.48  & 75.66$\pm$2.44 \\
& Hough      & 76.54$\pm$1.87  & 69.32$\pm$2.26 \\
& Probtrackx & 74.07$\pm$2.07  & 71.06$\pm$2.12 \\
& All        & 82.37$\pm$2.07  & 79.89$\pm$2.53 \\
\hline
\end{tabular}

\end{table}

\subsubsection{Effect of Model Depth}
\begin{figure}[H]
    \centering
    \includegraphics[width=0.6\linewidth]{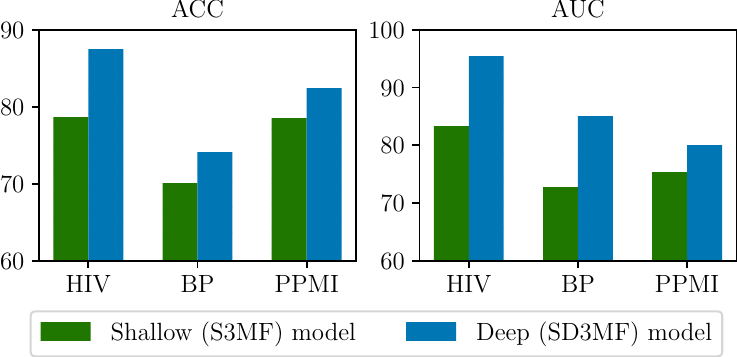}
    \caption{Shallow (S3MF) vs. deep (SD3MF) models in terms of ACC and AUC.}
    \label{fig:ablation}
\end{figure}

\autoref{fig:ablation} evaluates the effect of model depth by comparing the shallow S3MF variant with the deep SD3MF on the HIV, BP, and PPMI datasets in terms of ACC and AUC. Across all three datasets, SD3MF consistently achieves higher performance, suggesting that deep (multi-layer) factorization learns more informative community-level representations than a single-layer decomposition. {In our setting, deepness is an overparameterization~\cite{de2023consistent} that prevents converging to poor local minima~\cite{allen2019convergence}.}  
%\ngi{Also, the depth}
The improvements are particularly strong on HIV (78.60→87.50 ACC; 83.33→95.31 AUC) and BP (70.00→74.00 ACC; 72.73→85.05 AUC), where deeper modeling appears to better handle noise and capture cross-community interaction patterns relevant for diagnosis. On PPMI, SD3MF also yields steady gains (78.50→82.37 ACC; 75.30→79.89 AUC), indicating that depth remains beneficial even when single-model performance is already competitive. Overall, these results support the use of deep, hierarchical structure in SD3MF to capture multi-scale organization in brain networks and improve downstream classification.

% \end{figure}
% \begin{table}
% \centering
% % \footnotesize
% \scriptsize
% \caption{Evaluating the diagnostic accuracy of the SD3MF model across single-modal and multimodal neural connectivity data.}
% \label{tab:modals}
% \begin{tabular}{llccc}
% \hline
% \textbf{Dataset} & \textbf{Modality} & \textbf{Accuracy (\%)} & \textbf{AUC (\%)} \\
% \hline
% \multirow{3}{*}{HIV}
% & fMRI  & 77.16$\pm$3.22   & 89.58$\pm$4.89 \\
% & DTI   & 82.13$\pm$6.84   & 91.15$\pm$2.62 \\
% & Both  & 87.50$\pm$3.60   & 95.31$\pm$2.62 \\
% \hline
% \multirow{3}{*}{BP}
% & fMRI  & 71.00$\pm$5.50   & 78.18$\pm$2.50 \\
% & DTI   & 72.00$\pm$2.70   & 76.57$\pm$2.66 \\
% & Both  & 74.00$\pm$4.90   & 85.05$\pm$2.34 \\
% \hline
% \multirow{4}{*}{PPMI}
% & PICo       & 78.93$\pm$2.48  & 75.66$\pm$2.44 \\
% & Hough      & 76.54$\pm$1.87  & 69.32$\pm$2.26 \\
% & Probtrackx & 74.07$\pm$2.07  & 71.06$\pm$2.12 \\
% & All        & 82.37$\pm$2.07  & 79.89$\pm$2.53 \\
% \hline
% \end{tabular}
% \end{table}
% \begin{figure}
%     \centering
%     \includegraphics[width=0.4\linewidth]{ablation.pdf}
%     \caption{
%     Shallow (S3MF) vs.\ deep (SD3MF) models in terms of ACC and AUC on the HIV, BP, and PPMI datasets.
%     }
%     \label{fig:ablation}
% \end{figure}
\subsection{Interpretation of Learned Representations} 
We provide an intrinsic interpretability analysis by linking learned parameters to neurobiologically meaningful community structure and discriminative regional patterns. Specifically, we interpret (i) community organization and network composition from the membership matrices $\{\mathbf \Psi^{(m)}\}_{m=1}^{M}$ and subject-specific interaction matrices $\mathbf S_i$ of 2 groups, and (ii) salient ROIs from scores that combine $\{\mathbf\Psi^{(m)}\}_{m=1}^{M}$ with classifier coefficients. Both analyses are conducted in a modality-specific manner and under the weighted multimodal fusion setting to enable comprehensive interpretation across imaging modalities.
We illustrate these interpretations on the HIV dataset. 
% \todo{Nicolas: I don't understand (ii) well. You use the same info as in (i)? No use of the $S_i$'s? Can you clarify? Answer of Songlin: (i) is by both $\Psi$ and $S_i$ and (ii) is only by $\Psi$. Sorry for the confusion, and I have updated.} 

% \begin{figure}
%     \centering
%     \includegraphics[width=\linewidth]{model_interpretability/Communities.pdf}
%     \caption{Heatmap visualization of the learned basis matrices $\mathbf\Psi^{(m)}$ for DTI, fMRI, and multimodal settings. Each row corresponds to a latent community and each column to an ROI, with color intensity indicating the strength of ROI-to-community membership, illustrating how SD3MF organizes ROIs into modality-specific and multimodal community structures.}
%     \label{fig:sd3mf_communities}
% \end{figure}

\subsubsection{Interpreting Communities and Networks}
\begin{figure}
    \centering
    \includegraphics[width=1\linewidth]{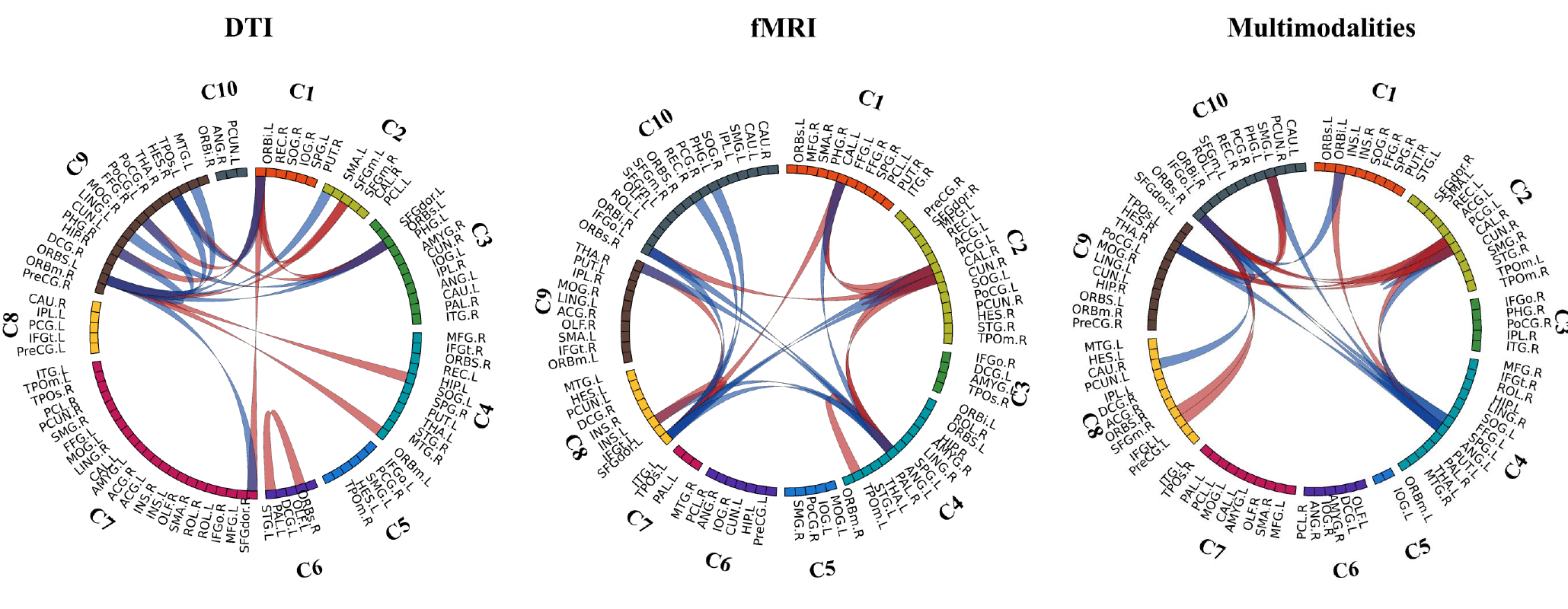}
    \caption{Interpretation of learned communities for the HIV cohort. ROIs are grouped by latent communities (C1 to C10), and ribbons indicate the relationship between ROIs, where red and blue denote relatively high and low loading values, respectively.}
    \label{fig:sd3mf_communities_chord}
\end{figure}

\begin{figure}[t]

        \centering
        \includegraphics[trim=20 7 20 7, clip, width=0.4\linewidth
        % , height = 2.3cm
        ]{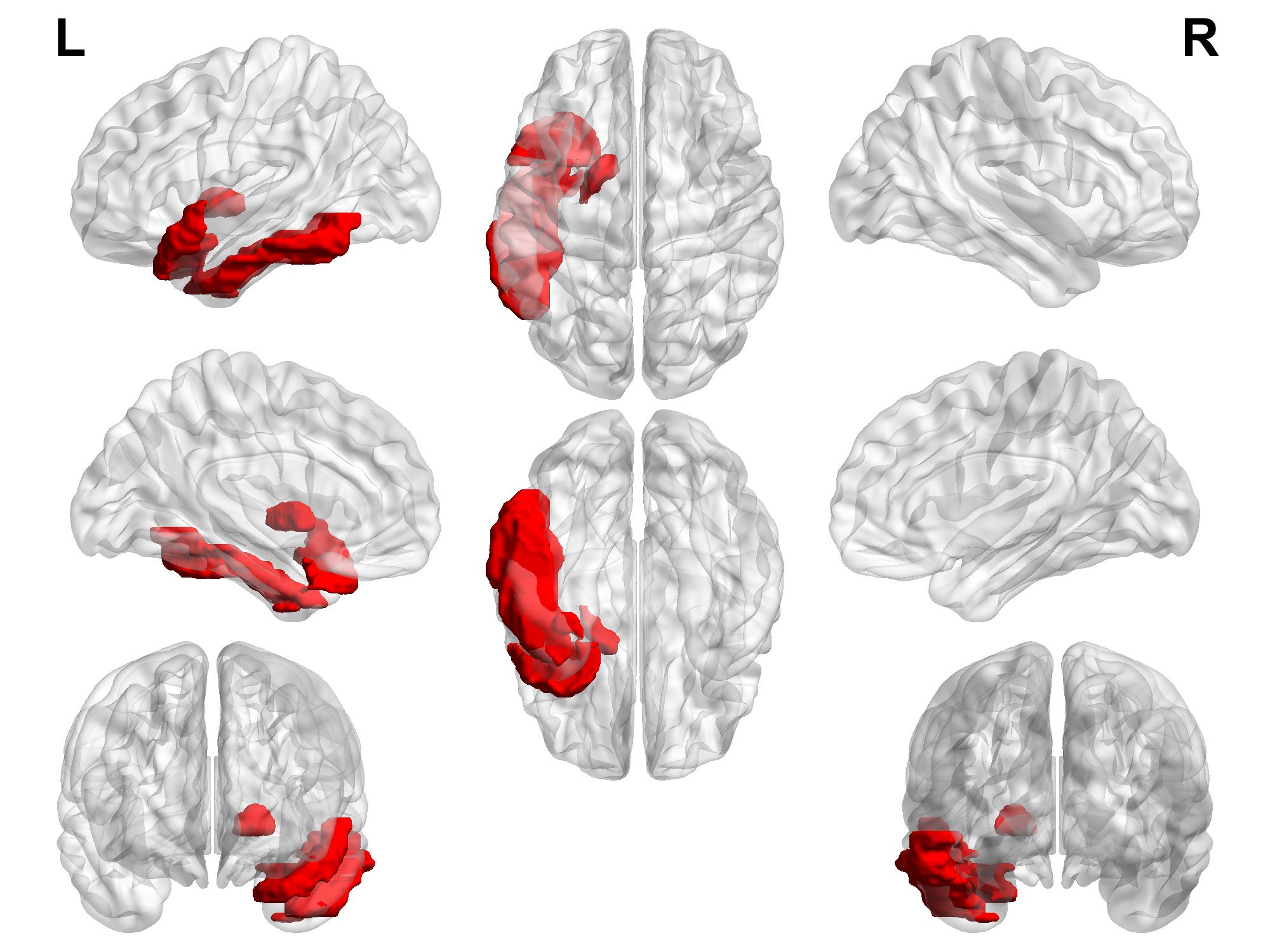}
        \caption{An fMRI-derived community (C7) by SD3MF on the HIV dataset: the left pallidum, left temporal pole, and left inferior temporal gyrus.}
        \label{fig:sd3mf_fMRI_community_7}
\end{figure}

We summarize the learned community composition on the HIV cohort for each modality (DTI, fMRI) and the multimodal setting using the corresponding basis matrix $\mathbf \Psi^{(m)}$. 
The number of communities is set to 10, aligning with established evidence that the brain functional architecture is organized into approximately ten primary representative networks \cite{smith2009correspondence}.
To make the community structure explicit, we assign each ROI $p$ to its dominant community via a hard membership rule $g^{(m)}(p)=\arg\max_{k}\Psi^{(m)}_{p,k}$, and report the dominant ROIs per community (C1, C2, ..., C10) accordingly, as shown in \autoref{fig:sd3mf_communities_chord}. Across modality settings, SD3MF consistently groups ROIs with coherent anatomical locations and functional roles into the same community. Multiple communities are enriched with occipital and early visual cortices, consistent with canonical visual resting-state networks \cite{yeo2011organization}. For example, \autoref{fig:sd3mf_fMRI_community_7} highlights an fMRI-derived community (C7 by fMRI) that jointly groups the pallidum with lateral temporal cortices (including the inferior temporal gyrus and temporal pole), consistent with prior evidence that HIV preferentially affects basal ganglia and temporal-limbic systems and is associated with disruptions spanning motor control and higher-order cognition \cite{israel2019different,thompson2015novel}.
 These patterns align with the modular and hierarchical organization of large-scale brain networks \cite{sporns2016modular} and provide a principled, interpretable partition of ROIs into communities that are directly optimized for the downstream classification objective.
To further facilitate qualitative inspection, \autoref{fig:sd3mf_communities_chord} provides a Circos visualization of ROI relationships, where ROIs are grouped by their dominant community. For each subject $i$ and modality $m$, we reconstruct the connectivity matrix as $\widehat{\mathbf A}_i^{(m)}=\mathbf\Psi^{(m)}\mathbf S_i\mathbf\Psi^{(m)\top}$. For each of the 4005 unique undirected connections among the 90 ROIs, we compare the reconstructed connection values of subjects with HIV and healthy controls using a two sided Welch $t$ test for independent samples. We apply the Benjamini and Hochberg false discovery rate (FDR) correction across all 4005 tests and retain connections with adjusted $p<0.05$. We then rank the retained connections by the absolute $t$ statistic and select the top 20 for visualization. The resulting chords show how connections between ROIs are distributed within and between communities. We observe modality-specific emphases that remain neurobiologically plausible: DTI-derived communities more prominently co-localize subcortical nuclei with frontal and temporal regions, consistent with the known sensitivity of diffusion MRI to fronto-subcortical and white-matter related alterations in HIV \cite{seider2016age}; in contrast, fMRI-derived communities exhibit clearer grouping of canonical cortical systems, consistent with resting-state fMRI capturing intrinsic functional network organization \cite{fox2007spontaneous,yeo2011organization}. Notably, the multimodal communities preserve these canonical motifs while integrating complementary ROIs across modalities, yielding communities that are simultaneously anatomically coherent and functionally interpretable~\cite{jin2020functional,marshall2015influence}.

\begin{figure}[t]
        \centering
        \includegraphics[trim=50 50 50 18, clip, width=0.9\linewidth]{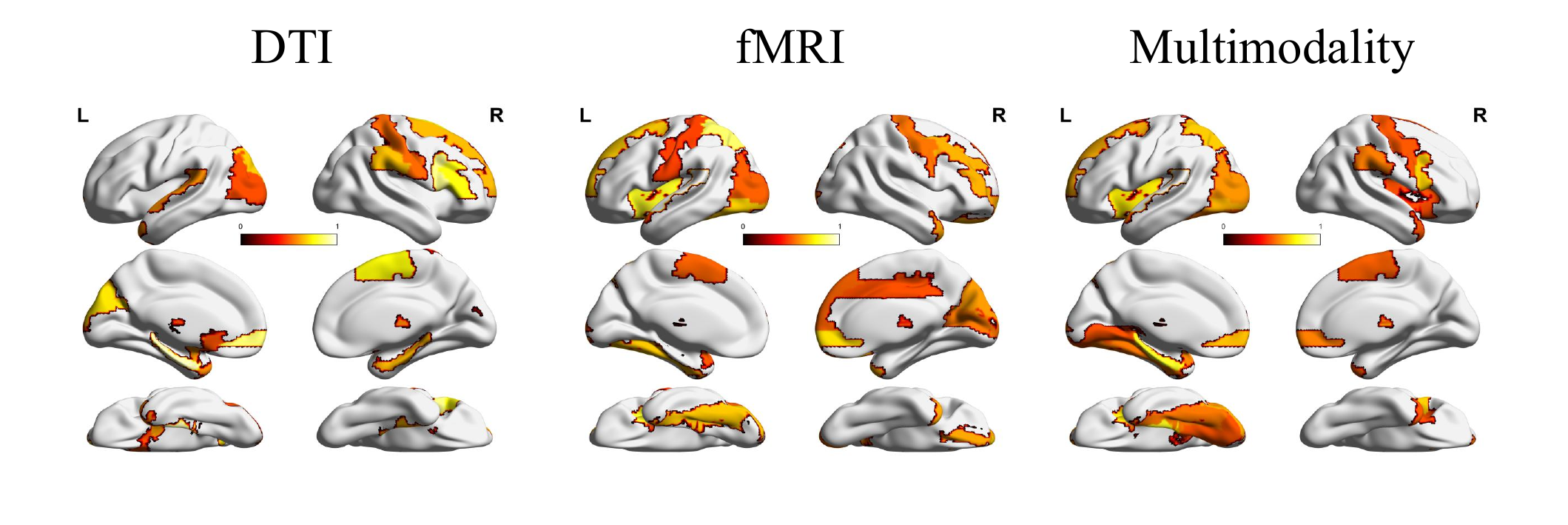}
        \caption{The top 20 ROIs ranked by classifier weighted membership scores for DTI, fMRI, and the multimodal setting. Results are rendered in lateral, medial, and ventral views.}
        \label{fig:sd3mf_SalientROI}
    \vspace{-10pt} % Standard NeurIPS space-saving trick
\end{figure}

% \begin{figure}
%     \centering
%     \includegraphics[width=0.35\linewidth
%     % , height = 3cm
%     ]{model_interpretability/fMRI_community_7.png}
%     \caption{An fMRI-derived community (C7) by SD3MF {on the HIV dataset}: the left pallidum, left temporal pole, and left inferior temporal gyrus.}
%     \label{fig:sd3mf_fMRI_community_7}
%     \vspace{-10pt}
% \end{figure}

\subsubsection{Interpreting Salient ROIs}
To identify salient ROIs for distinguishing HC and HIV, we combine the learned membership matrices with the magnitudes of the classifier coefficients. Let $\mathbf B\in\mathbb R^{K\times K}$ be the matrix obtained by reshaping the classifier coefficient vector stored as \texttt{beta[1]} in the implementation, with $K=10$. The importance of community $k$ and the saliency score of ROI $p$ in modality $m$ are defined as
\begin{align}
c_k &= \sum_{j=1}^{K}|B_{kj}|+\sum_{j=1}^{K}|B_{jk}|, \label{eq:community_saliency}\\
s_p^{(m)} &= \sum_{k=1}^{K}\Psi_{pk}^{(m)}c_k. \label{eq:roi_saliency}
\end{align}
Each membership loading is therefore weighted by the importance of its community to the classifier. The scores are computed directly from the learned model parameters. Before normalization, the multimodal score is obtained by combining the modality scores as
\begin{equation}\label{eq:multimodal_roi_saliency}
s_p^{\mathrm{multi}}=0.49s_p^{\mathrm{fMRI}}+0.51s_p^{\mathrm{DTI}}.
\end{equation}
For DTI, fMRI, and the multimodal setting separately, we apply min max normalization across all 90 ROIs to map the scores to $[0,1]$, rank them in descending order, and select the top 20. As shown in \autoref{fig:sd3mf_SalientROI}, BrainNet Viewer~\cite{xia2013brainnet} renders the selected ROIs in lateral, medial, and ventral views for comparison across settings.

We found that the salient ROIs for distinguishing HIV and HC span fronto-executive cortices (e.g., SFGdor, MFG, IFGoperc/IFGtriang, and orbitofrontal regions), salience/limbic-temporal nodes (insula, PHG, STG, and amygdala), subcortical nuclei, and visual cortices. This pattern is largely consistent with prior HIV neuroimaging evidence showing preferential involvement of frontal and cingulate/insula regions as well as limbic structures in HIV, which has been linked to cognitive decline \cite{xu2023meta,israel2019different}. Modality-specific findings were also in line with established modality sensitivities: DTI primarily highlighted fronto-subcortical circuits and basal ganglia/thalamic structures, consistent with reports of HIV-associated white-matter and fronto-subcortical integrity loss \cite{seider2016age}, whereas fMRI emphasized cortical network nodes consistent with dissociated functional abnormalities \cite{li2025divergent}. Under our  setting, these ROIs were jointly confirmed and strengthened across modalities, and the additional emphasis on visual-network regions is supported by HIV-related findings in visual cortices \cite{li2025divergent}.

\subsubsection{Learned Membership Matrix}
\begin{figure}
    \centering
    \includegraphics[width=0.9\linewidth]{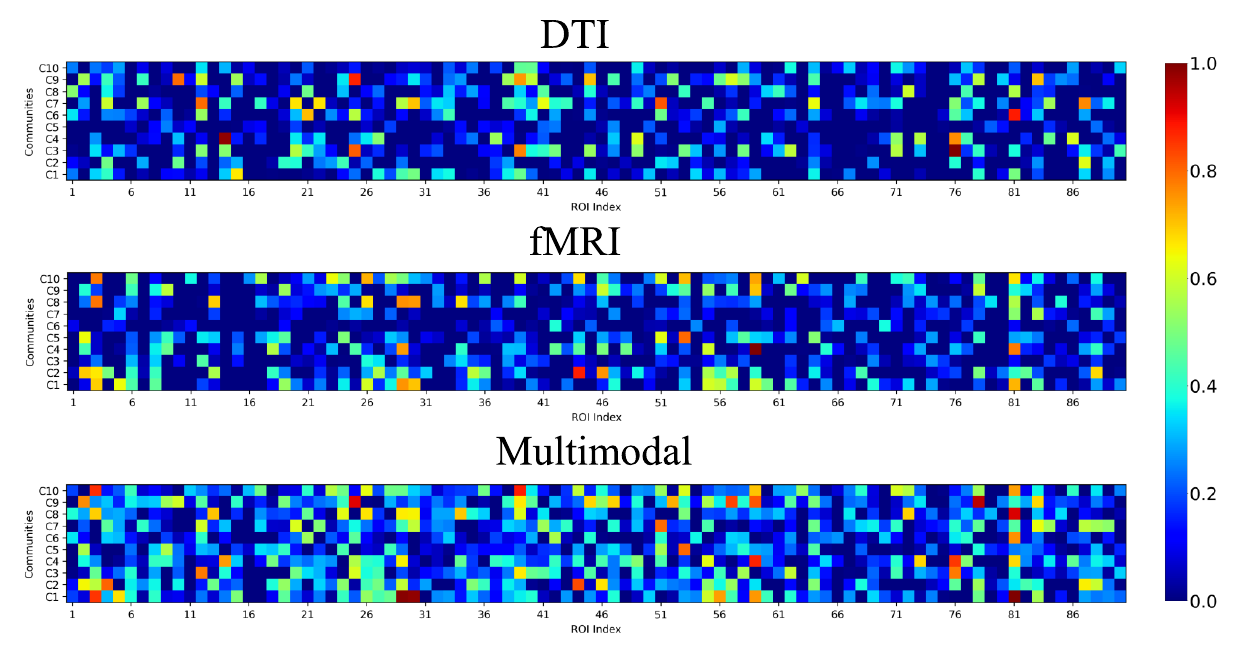}
    \caption{Visualization of the learned membership matrices $\mathbf\Psi^{(m)}$ for DTI, fMRI, and Multimodal settings on the HIV dataset.}
    \label{fig:sd3mf_communities}
\end{figure}

\autoref{fig:sd3mf_communities} visualizes the learned membership matrices $\mathbf{\Psi}^{(m)}$ for DTI, fMRI, and the multimodal setting on the HIV cohort. Each row corresponds to a latent community, and each column corresponds to an ROI, with color intensity indicating the strength of ROI-to-community membership. This representation provides a compact view of how SD3MF organizes ROIs into modality-specific and multimodal community structures. We observe that the learned memberships exhibit sparse and structured activation patterns, where subsets of ROIs consistently show strong loadings within specific communities, suggesting the emergence of coherent functional and anatomical modules rather than diffuse or uniform assignments. Compared to single-modality settings, the multimodal membership matrix shows denser and more balanced activation across communities, indicating complementary integration of modality-specific information and enhanced community expressiveness. These patterns support that the proposed hierarchical factorization captures meaningful modular organization of the brain connectome and provides an interpretable latent representation that is subsequently leveraged for downstream classification.

\begin{figure}
    \centering
    \includegraphics[width=0.8\linewidth]{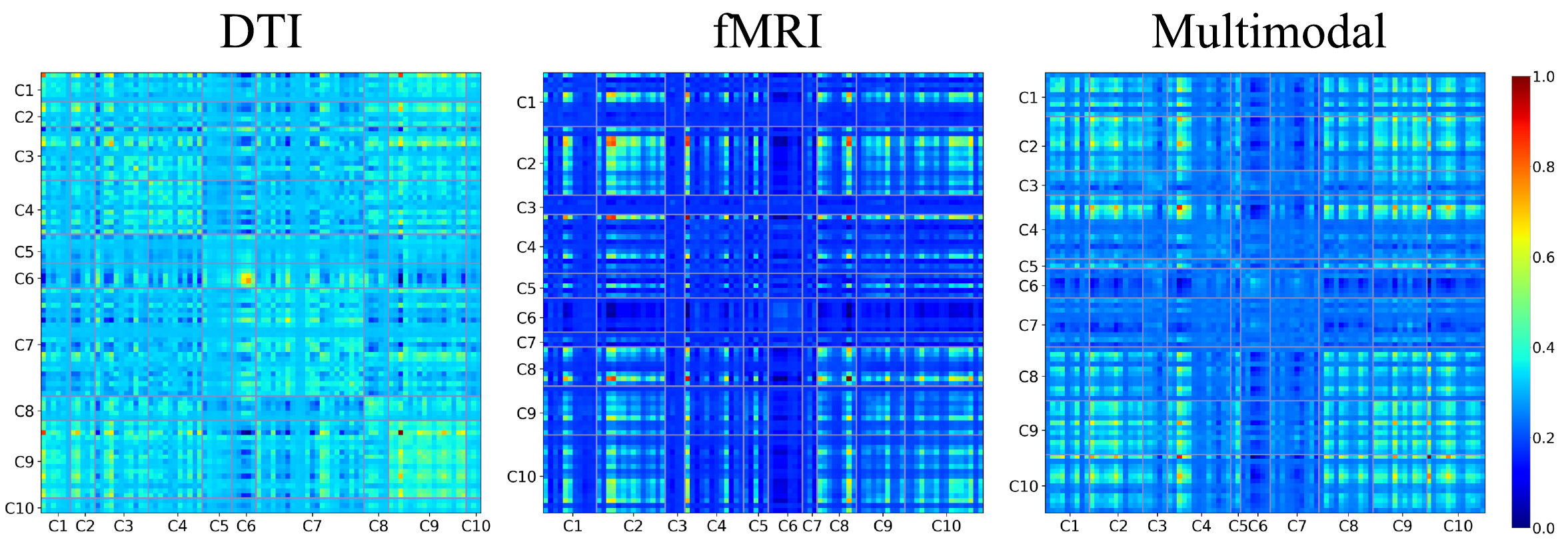}
    \caption{Approximation matrices $\widehat{\mathbf A}_{\mathrm{HIV}}^{(m)}=\mathbf\Psi^{(m)}\bar{\mathbf S}_{\mathrm{HIV}}\mathbf\Psi^{(m)\top}$ in ROI space for the HIV cohort under DTI, fMRI, and multimodal settings. Entries represent approximated ROI-to-ROI connectivity strengths. ROIs are reordered by dominant community assignments (C1--C10) shown on the axes for visualization.}
    \label{fig:sd3mf_approximation}
\end{figure}

\subsubsection{Connectivity Analysis via Approximation Matrix}
To inspect the ROI-to-ROI connectivity captured by SD3MF on the HIV dataset, we visualize the approximation matrices reconstructed in the original ROI space. For each modality in HIV group, we compute the group-averaged interaction matrix $\bar{\mathbf S}_{\mathrm{HIV}}$ and reconstruct the connectivity approximation as
\begin{equation}
\widehat{\mathbf A}^{(m)}_{\mathrm{HIV}}=\mathbf \Psi^{(m)}\,\bar{\mathbf S}_{\mathrm{HIV}}\,\mathbf \Psi^{(m)\top}.
\end{equation}
where $\widehat{\mathbf A}^{(m)}_{\mathrm{HIV}}$ represents the SD3MF-induced approximation of HIV-related ROI connectivity. In \autoref{fig:sd3mf_approximation}, ROIs are reordered according to their dominant community assignments (shown on the axes) for visualization.
As shown in \autoref{fig:sd3mf_approximation}, the approximation matrices exhibit structured block-like patterns rather than diffuse connectivity, indicating coherent subnetwork organization captured by the model. The DTI approximation shows smoother and more globally distributed patterns, while the fMRI approximation presents more heterogeneous and localized structures. The multimodal result integrates both characteristics and yields more coherent connectivity patterns, demonstrating the benefit of combining complementary structural and functional information.

% \begin{figure}
%     \centering
%     \includegraphics[trim=50 50 50 18, clip, width=0.5\linewidth]{model_interpretability/SalientROI.pdf}
%     \caption{Surface visualization of modality-specific and multimodal salient ROIs for distinguishing HIV from healthy controls. Results are rendered on cortical surfaces in lateral, medial, and ventral views for the left (L) and right (R) hemispheres.}
%     \label{fig:sd3mf_SalientROI}
% \end{figure}

\subsection{Empirical convergence} \label{sec:empiricalconv}

\begin{figure}[H]
    \centering
    \begin{subfigure}{0.32\linewidth}
        \centering
        \includegraphics[width=\linewidth]{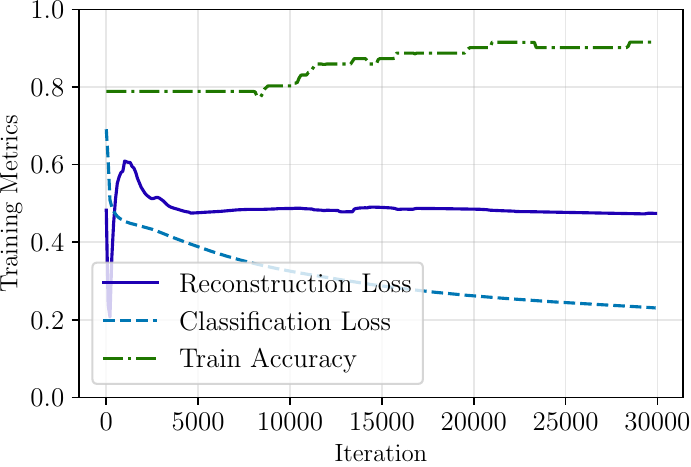}
        \caption{HIV}
        \label{fig:conv_hiv}
    \end{subfigure}
    \hfill
    \begin{subfigure}{0.32\linewidth}
        \centering
        \includegraphics[width=\linewidth]{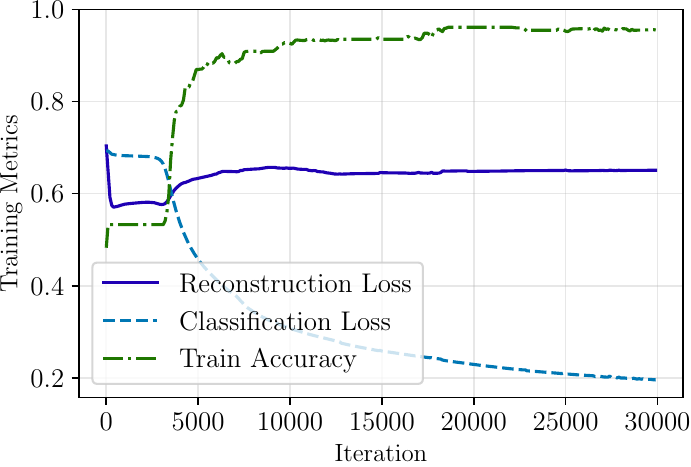}
        \caption{BP}
        \label{fig:conv_bp}
    \end{subfigure}
    \hfill
    \begin{subfigure}{0.32\linewidth}
        \centering
        \includegraphics[width=\linewidth]{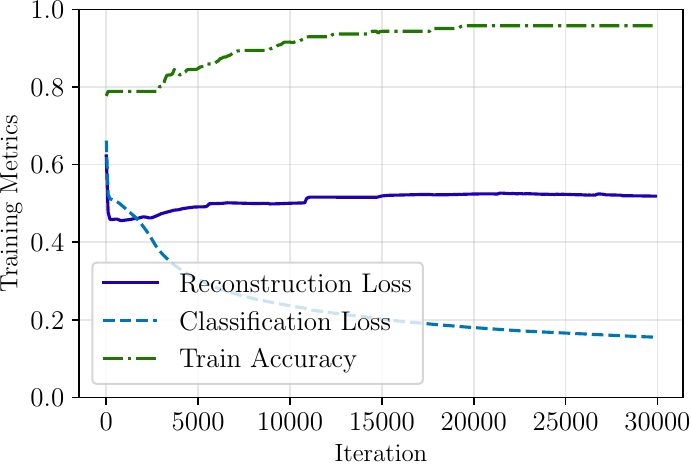}
        \caption{PPMI}
        \label{fig:conv_ppmi}
    \end{subfigure}

    \caption{Convergence behavior of the proposed SD3MF with learning rate $10^{-5}$ on three datasets showing reconstruction loss, classification loss, and training accuracy.}
    \label{fig:conv_all}
\end{figure}

\autoref{fig:conv_all} shows the evolution of the training metrics of SD3MF on the HIV, BP, and PPMI datasets. 
 SD3MF exhibits stable convergence under SGD with a small learning rate ($10^{-5}$) and small-scale (near-zero) initialization, which provides implicit regularization by biasing deep MF toward low-rank, well-conditioned solutions without explicit penalties. During training, the classification loss decreases steadily and training accuracy rises and then plateaus, indicating reliable optimization of the supervised objective. The reconstruction loss initially drops but later shows a mild increase, consistent with the joint objective: as the model becomes more label-discriminative, it can sacrifice reconstruction fidelity to better align the learned community interactions with the classification task. Overall, the curves suggest controlled, non-divergent training dynamics where small initialization and gentle updates regularize learning while the optimizer navigates the reconstruction–prediction balance.

\section{Conclusion}\label{sec:conclusion}
{
We proposed SD3MF, a supervised deep multimodal matrix factorization framework that generalizes SNMTF from unsupervised graph representation to deep, end-to-end multimodal brain network classification. By combining hierarchical matrix factorization, subject-specific community interaction representations, supervised prediction, and adaptive multimodal fusion within a unified model, SD3MF provides a structured alternative to conventional CNN- and GNN-based approaches. In particular, the learned low-dimensional representations retain explicit community-level information, allowing the predictive patterns associated with clinical or cognitive labels to be interpreted in terms of interactions among latent brain communities. Experimental results on multiple multimodal connectome datasets demonstrate that SD3MF achieves strong predictive performance compared with tensor-based, CNN, and GNN baselines, while generally exhibiting lower variability across repeated runs. The ablation studies further show that multimodal fusion, model depth, and the balance between reconstruction and supervised objectives each contribute to the overall performance. Beyond predictive accuracy, the learned community structures and subject-specific interaction matrices provide an interpretable view of the network patterns underlying the classification decisions, highlighting the potential of structured matrix factorization as an effective bridge between predictive performance and interpretability in multimodal network learning. 

Despite these advantages, the current framework has several directions for further development. The use of a linear supervised head reflects a deliberate trade-off between interpretability and modeling flexibility; future work could investigate more expressive yet interpretable prediction mechanisms, such as Kolmogorov-Arnold Networks (KAN)~\cite{liu2025kan}. Extending SD3MF to temporal or longitudinal multimodal networks would also enable the evolution of latent community structures and their interactions to be modeled over time. Another promising direction is to investigate more flexible fusion mechanisms for handling heterogeneous or partially missing modalities. More broadly, although this work focuses on brain connectomes, the proposed formulation is applicable to other multimodal network-valued data, including biomedical, social, sensor, and knowledge networks.
}

\section*{Acknowledgments}
  
Amjad Seyedi and Nicolas Gillis gratefully acknowledge support from the European Union (ERC Consolidator Grant, eLinoR, no.\ 101085607). Lifang He and Akwum Onwunta gratefully acknowledge support from the U.S. Department of Energy (DOE grant DE-SC0025801, "Harnessing Nonnegative Matrix Factorization for Advanced Computational Materials Modeling").
\bibliography{references}

\end{document}